\documentclass[runningheads]{llncs}

\usepackage{eccv} 

\usepackage{eccvabbrv}

\usepackage{graphicx}
\usepackage{booktabs}

\usepackage[accsupp]{axessibility}  

\usepackage{hyperref}

\usepackage{orcidlink}

\newcommand{\smalltitle}[1]{ \vspace{1mm}{\noindent\textbf{#1}\hspace{1mm}}}

\usepackage{wrapfig}
\usepackage{algorithm}
\usepackage{algpseudocode}

\usepackage{subcaption}
\usepackage{caption}

\floatstyle{ruled}
\restylefloat{algorithm}

\usepackage{booktabs}
\usepackage[table]{xcolor}
\usepackage{array}
\usepackage{graphicx} 
\newcolumntype{L}[1]{>{\raggedright\arraybackslash}p{#1}}
\usepackage{makecell}
\usepackage{tabularx}
\usepackage{array}
\newcolumntype{Y}{>{\centering\arraybackslash}X} 
\definecolor{mygray}{gray}{0.4}
\definecolor{lightgreen}{RGB}{180,255,180}
\usepackage{multirow}
\usepackage{needspace}

\begin{document}

\title{Moving Beyond Diversity: Visual Token Pruning as Subspace Reconstruction for Efficient VLMs}

\titlerunning{SPARE}

\author{Jaeyeon Lee, Shunjie Wen \and Dong-Wan Choi}
\authorrunning{J.~Lee et al.}

\institute{Princeton University, Princeton NJ 08544, USA \and
Springer Heidelberg, Tiergartenstr.~17, 69121 Heidelberg, Germany
\email{lncs@springer.com}\\
\url{http://www.springer.com/gp/computer-science/lncs} \and
ABC Institute, Rupert-Karls-University Heidelberg, Heidelberg, Germany\\
\email{\{abc,lncs\}@uni-heidelberg.de}}

\institute{Inha University\\
\email{\{dlwodus159, wenshunjie\}@inha.edu, dchoi@inha.ac.kr}}

\maketitle


\begin{abstract}
Despite their remarkable performance, Vision Language Models (VLMs) incur substantial computational overhead due to the large number of visual tokens. While diversity maximization has become a dominant strategy for token reduction, existing methods rely on cosine-based normalized similarity that discards magnitude information, failing to faithfully approximate the original feature representation and leading to suboptimal performance, particularly on compositional multi-skill reasoning tasks.
In this paper, we introduce SPARE, a sub\underline{SPA}ce \underline{RE}construction method that reformulates token pruning as a \textit{column subset selection problem} and explicitly minimizes reconstruction error. By iteratively selecting tokens with large projection residuals, SPARE performs reconstruction-driven pruning beyond angular diversity.
Moreover, we reveal a counterintuitive anti-relevance phenomenon: tokens with lower image-text relevance score can better preserve contextual information. Based on this finding, we incorporate anti-relevance into SPARE as an additional selection criterion to promote context-aware token selection.
Extensive experiments across multiple VLMs and benchmarks demonstrate that SPARE consistently achieves state-of-the-art performance, with strong gains on compositional tasks. When applied to LLaVA, SPARE removes up to 94\% of visual tokens while retaining 95\% of the baseline performance, all in a fully training-free manner.

\end{abstract}

\begin{figure}[t]
  \centering
  \begin{subfigure}{0.62\textwidth}
    \centering
    \includegraphics[width=\linewidth]{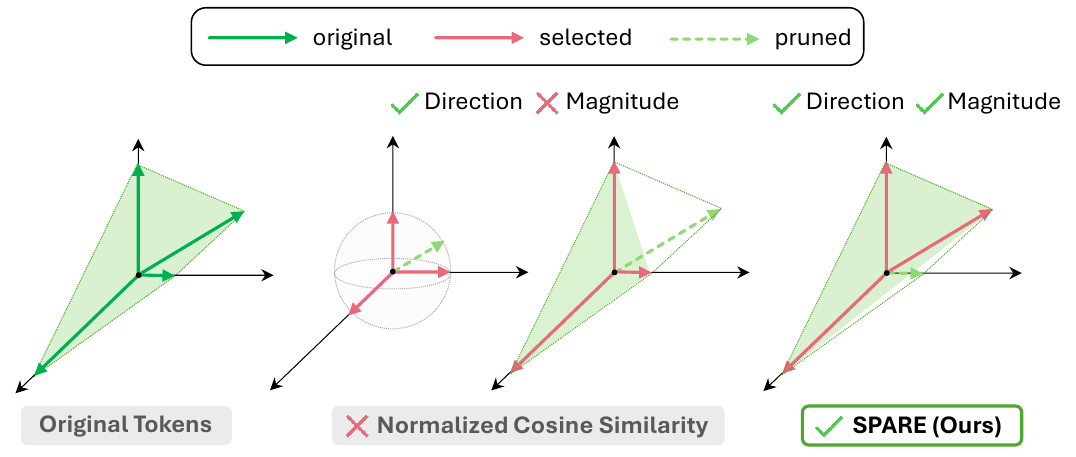}
    \caption{Comparison of Representation Preservation}
    \label{fig1_a_spacedifference}
  \end{subfigure}
  \hspace{-1.5mm}
  \begin{subfigure}{0.37\textwidth}
    \centering
    \includegraphics[width=\linewidth]{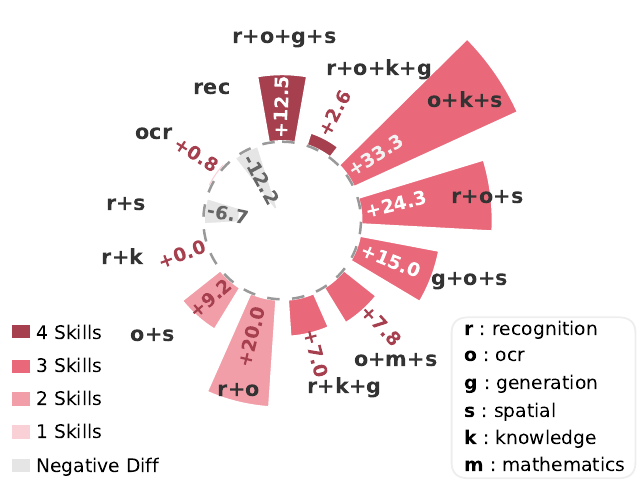}
    \caption{Multi-Skill Performance Gap}
    \label{fig1_b_multi-skill}
  \end{subfigure}
  \caption{\textbf{Diversity-Based Token Selection vs. SPARE (Ours).}
  (a) Each visual token is represented as a vector in the embedding space. Diversity-based methods (\eg, CDPruner \cite{zhang2025beyond}) select tokens based on normalized angular dispersion, whereas SPARE performs subspace reconstruction that accounts for both direction and magnitude. (b) Performance gap (SPARE -- CDPruner) on multi-skill task combinations in MM-Vet \cite{YuYLWL0WW24} using LLaVA-1.5-7B \cite{LiuLLL24} under visual token pruning.}
  \label{fig:overall_intro}
  \vspace{-2mm}
\end{figure}

\section{Introduction}
Building upon the success of Large Language Models (LLMs) \cite{abs-2310-06825, abs-2302-13971, vicuna2023}, recent works have increasingly extended them to the visual domain, giving rise to modern Vision Language Models (VLMs) \cite{abs-2502-13923, abs-2504-10479, LiuLWL23a, 0080ZGZ00ZZL0L25}. A common design directly incorporates visual tokens into the LLM input sequence alongside textual tokens, achieving strong performance across diverse vision-language benchmarks. However, due to the spatial nature of images, visual inputs can produce thousands or even tens of thousands of tokens, especially for high-resolution images or videos, substantially increasing the computational cost of VLMs. To alleviate this bottleneck, numerous approaches aim to reduce the number of visual tokens in VLMs.


Recent progress in visual token reduction falls into two paradigms: attention-based and diversity-based approaches. Attention-based methods \cite{0020FMZ0CGONKZ25, abs-2410-17247, ChenZLBLZC24} select tokens using attention scores from vision encoders or LLMs, whereas diversity-based methods \cite{zhang2025beyond, abs-2509-24837, WenGWZZLHZ25} reduce redundancy by minimizing similarity among selected tokens. Since attention-based reduction can underperform even random pruning \cite{WenGLH025, WenGWZZLHZ25} and is often incompatible with efficient attention operators \cite{DaoFERR22}, diversity-based strategies become the predominant direction in recent research due to their strong performance.



However, existing diversity-based methods \cite{AlvarSAZ25,zhang2025beyond, WenGWZZLHZ25} still remain limited because they rely on cosine-based normalized similarity.
By treating token pruning as a dispersion problem, these approaches deliberately discard token magnitudes through normalization, rather than explicitly approximating the original feature representation. As a result, the selected tokens may deviate substantially from the overall representation formed by the full token set (see Fig. \ref{fig1_a_spacedifference}), leading to larger reconstruction errors from the original feature matrix (see Table~\ref{tab6_frobenius_error}). As presented in Fig.~\ref{fig1_b_multi-skill}, this limitation translates into suboptimal performance, particularly on challenging reasoning tasks that require multi-skill compositional understanding (\eg, \textit{``What is the number displayed on the motorcycle on the right?''}, requiring all of recognition, OCR and spatial reasoning skills).



To address this, we propose SPARE, a sub\underline{SPA}ce \underline{RE}construction token pruning method that reformulates visual token pruning from the perspective of the \textit{column subset selection problem} (CSSP)~\cite{BoutsidisMD09, FarahatEGK15}.
Instead of merely encouraging inter-token dispersion based on normalized similarity, SPARE explicitly aims to approximate the original feature matrix using a subset of tokens.
To this end, SPARE performs greedy subspace reconstruction by iteratively selecting the token with the largest projection error (\ie, residual norm) relative to the current subset, a strategy grounded in the \textit{rank-revealing QR} (RRQR) principle~\cite{chan1987rank}.
Since the residual norm inherently captures both directional and magnitude information, SPARE preserves the original feature representation more faithfully (see Fig. \ref{fig1_a_spacedifference}) than diversity-based approaches operating in a normalized angular space.

Beyond faithful visual reconstruction, we further observe an \textit{anti-relevance phenomenon} in visual token selection with respect to the textual query, \ie, retaining tokens with lower image-text relevance score can yield better downstream performance. This is contrary to the common belief among prior approaches~\cite{SongWCWGW25, zhang2025beyond} that tokens with higher image-text relevance, as measured by CLIP \cite{RadfordKHRGASAM21}, are more important for preserving task-relevant information. To examine this assumption, we explicitly quantify image-text relevance and analyze its influence on token selection and downstream performance. Our analysis reveals a critical mismatch: maximizing such relevance does not guarantee the preservation of task-relevant information. Even slight pruning biased toward low-relevance tokens (\ie, retaining most high-relevance ones) can cause substantial performance degradation on certain benchmarks. Motivated by this finding, we incorporate anti-relevance as an additional guiding principle within our SPARE method.

Grounded in this design, we summarize our contributions as follows. (1) This is the first work to reformulate visual token pruning as a CSSP, providing an explicit approximation framework for token selection in VLMs. In particular, we explicitly consider the token's magnitude during selection, moving beyond normalized similarity and enabling reconstruction-based representation preservation. (2) We systematically analyze the role of image-text relevance and reveal an anti-relevance phenomenon, demonstrating the effectiveness of anti-relevance guidance in token selection. (3) SPARE is training-free and plug-and-play, retaining 95\% of the baseline performance even after pruning up to 94\% of visual tokens, and consistently outperforming prior methods across benchmarks, particularly on challenging multi-skill compositional tasks (see Fig. \ref{fig1_b_multi-skill}).

\section{Related Works}
\label{sec:related}

Modern Vision Language Models (VLMs) increasingly adopt high-resolution and multi-frame inputs, resulting in thousands or even tens of thousands of visual tokens per sample~\cite{liu2024llavanext, abs-2504-10479, abs-2407-07895}. For example, LLaVA-NeXT~\cite{liu2024llavanext} processes up to 2,880 visual tokens for high-resolution images, while video-capable models such as InternVL~\cite{abs-2504-10479} and LLaVA-NeXT-Interleave~\cite{abs-2407-07895} can involve tens of thousands of tokens due to temporal accumulation. This growth in token counts has motivated extensive research on improving the efficiency of VLMs. Existing token reduction approaches can be broadly categorized into two branches: attention-based and diversity-based strategies.

\smalltitle{Attention-Based Visual Token Reduction in VLMs.}
This line of work evaluates token importance individually using attention scores derived from vision encoders or LLMs, pruning tokens deemed less influential based on their attention values. For example, FastV~\cite{ChenZLBLZC24} prunes tokens according to their average attention scores after the shallow layers, while SparseVLM~\cite{0020FMZ0CGONKZ25} and PDrop~\cite{abs-2410-17247} employ image-text attention for multi-stage token pruning within LLM layers.

However, recent studies~\cite{WenGLH025, WenGWZZLHZ25} report that attention-based pruning can perform even worse than random pruning under certain settings. Moreover, because token selection relies on intermediate attention extraction, it complicates integration with efficient attention implementations such as FlashAttention~\cite{DaoFERR22}.

\smalltitle{Diversity-Based Visual Token Reduction in VLMs.}
In contrast to attention-based methods that assess token importance individually, diversity-based approaches operate at the set level, minimizing redundancy by encouraging mutual dissimilarity among selected tokens. DivPrune~\cite{AlvarSAZ25} formulates token selection as a max-min diversity problem to greedily select the most distant tokens, while DART~\cite{WenGWZZLHZ25} first identifies pivot tokens before expanding to diverse neighbors. Similarly, CDPruner~\cite{zhang2025beyond} employs a determinantal point process to maximize statistical diversity among selected tokens.

Compared to attention-based methods, diversity-based approaches have shown stronger empirical performance and dominate recent state-of-the-art token reduction results. However, because their underlying view frames token selection as a dispersion problem rather than an approximation problem, they naturally rely on normalized similarity metrics. Consequently, they do not explicitly preserve the original feature subspace structure. This limitation motivates us to reformulate visual token pruning as a column subset selection problem (CSSP) and address it from a reconstruction perspective.

\smalltitle{Column Subset Selection Problem.}
The column subset selection problem (CSSP)~\cite{BoutsidisMD09} is a fundamental matrix approximation problem that seeks a subset of columns minimizing the reconstruction error of the original matrix. Existing solution algorithms are either randomized~\cite{DrineasMM08,DeshpandeRVW06} or deterministic~\cite{GuE96, chan1987rank}.
In this work, we adopt a deterministic approach based on rank-revealing QR (RRQR) factorization~\cite{chan1987rank} to ensure consistent and reproducible token selection. Specifically, RRQR permutes columns according to their contribution to the dominant structure of the feature matrix, yielding a stable and representative subset.

\section{Method}
We first formalize the visual token pruning problem within the VLM pipeline, and then present our SPARE method in detail. The conceptual overview of SPARE is illustrated in Fig.~\ref{fig2_overview}.

\subsection{Problem Formulation}
\begin{figure}[t]
    \centering
    \includegraphics[width=0.95\linewidth]{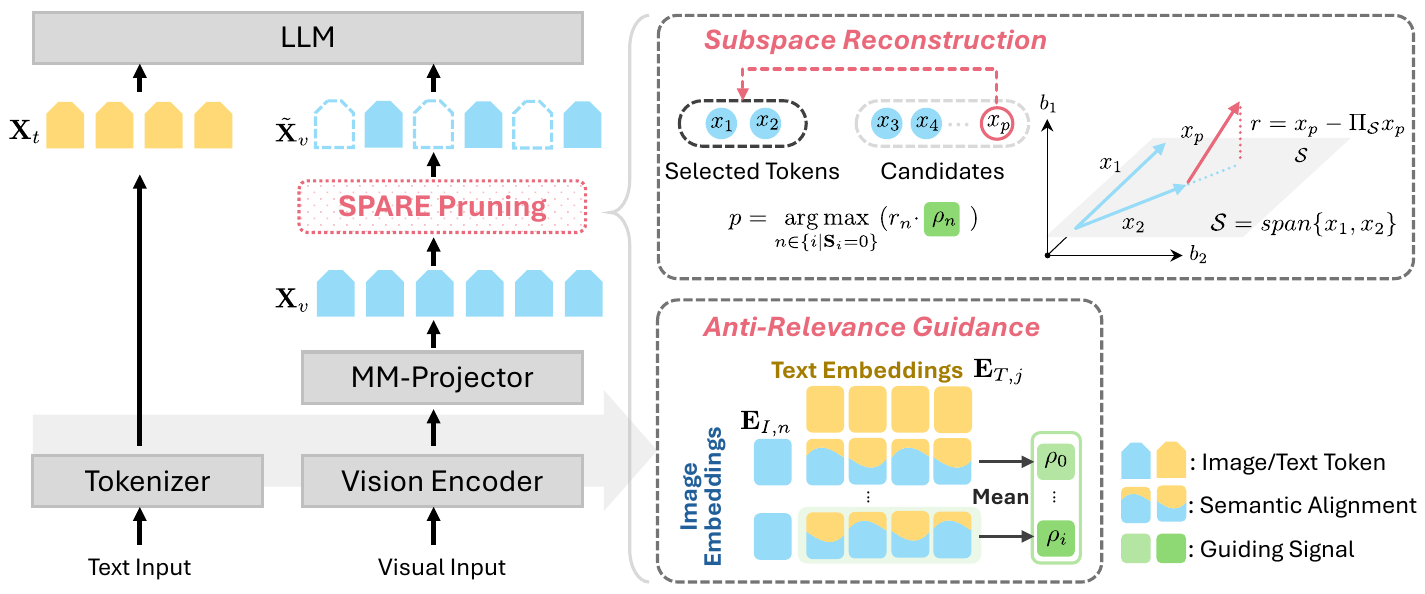}
    \caption{\textbf{Overview of SPARE.} SPARE approximates the visual subspace by iteratively selecting tokens based on reconstruction residuals, while incorporating anti-relevance guidance derived from image--text semantic alignment to preserve complementary contextual information.}
    \label{fig2_overview}
    \vspace{-2mm}
\end{figure}

\smalltitle{Vision Language Model Pipeline.}
A typical modern VLM consists of a vision encoder, a multimodal projector, and an LLM, as shown in the left panel of Fig.~\ref{fig2_overview}.
Given an image $I$, the vision encoder followed by the multimodal projector produces a sequence of visual tokens $\mathbf{X}_v = [\mathbf{x}_1, \dots, \mathbf{x}_N] \in \mathbb{R}^{N \times d}$, where $N$ denotes the number of visual tokens and $d$ is the embedding dimension of the LLM.
Given a text input $T$, the tokenizer produces textual token embeddings 
$\mathbf{X}_t = [\mathbf{t}_1, \dots, \mathbf{t}_M] \in \mathbb{R}^{M \times d}$, 
where $M$ denotes the number of textual tokens.
The visual and textual token embeddings are concatenated and fed into the LLM as a multimodal input sequence.
The LLM then models the conditional probability of an output sequence $Y = (y_1, \dots, y_\ell)$ as:
\begin{equation}
\label{eq1}
P(Y \mid I, T) = \prod_{i=1}^{\ell} P(y_i \mid y_{<i}, I, T),
\end{equation}
where $\ell$ is the number of output tokens. 

In modern VLMs, the number of visual tokens is typically much larger than that of textual tokens (\ie., $N \gg M$), leading to significant computational and memory overhead in the LLM. Therefore, reducing the number of visual tokens while preserving task-relevant information becomes a critical problem.

\smalltitle{Visual Token Pruning Problem.}
Given visual token embeddings $\mathbf{X}_v \in \mathbb{R}^{N \times d}$ and textual token embeddings $\mathbf{X}_t \in \mathbb{R}^{M \times d}$, 
the goal of visual token pruning is to select a subset of visual tokens that preserves the output behavior of the VLM. 
More specifically, we seek a subset of indices $\mathcal{S} \subset \{1, \dots, N\}$ with $|\mathcal{S}| = k \ll N$, 
and denote by $\tilde{\mathbf{X}}_v \in \mathbb{R}^{k \times d}$ the visual token embeddings corresponding to the selected indices $\mathcal{S}$. Let $\mathcal{F}(\mathbf{X}_v, \mathbf{X}_t)$ denote the output distribution of the LLM given visual and textual embeddings. Then, the ideal objective of visual token pruning can be formulated as:
\begin{equation}
\label{eq:loss:ideal}
\min_{\mathcal{S} \subset \{1,\dots,N\}} 
\; \mathcal{L}\Big(
\mathcal{F}(\tilde{\mathbf{X}}_v, \mathbf{X}_t),
\mathcal{F}(\mathbf{X}_v, \mathbf{X}_t)
\Big)
\quad \text{s.t. } |\mathcal{S}| = k.
\end{equation}
where $\mathcal{L}(\cdot,\cdot)$ measures the discrepancy between the two output distributions.

However, directly minimizing Eq.~\eqref{eq:loss:ideal} is intractable in dynamic pruning, as pruning is performed prior to LLM inference and evaluating the discrepancy requires executing the full autoregressive process. Consequently, existing token pruning methods rely on heuristic strategies. Some estimate token importance based on attention scores, while others encourage diversity among selected tokens to better cover the visual feature space. Nevertheless, the fundamental difficulty discussed in Section~\ref{sec:related} still remains.

\subsection{Visual Token Selection via Subspace Reconstruction}
\label{sec:recon}

\smalltitle{Token Selection as Reconstruction.}
To address the fundamental challenge above, we reinterpret visual token pruning as a reconstruction problem at the input level. Rather than relying on importance or diversity measures, we aim to directly approximate the original visual embeddings as follows:
\begin{equation}
\label{eq:loss:approx}
\min_{\mathcal{S} \subset \{1,\dots,N\}} 
\; \mathcal{L}\big(
\tilde{\mathbf{X}}_v, \mathbf{X}_v
\big)
\quad \text{s.t. } |\mathcal{S}| = k.
\end{equation}
Thus, Eq.~\eqref{eq:loss:approx} provides a tractable objective for approximating the ideal formulation in Eq.~\eqref{eq:loss:ideal}.

\smalltitle{Reconstruction as CSSP.}
More concretely, we formalize the reconstruction loss in Eq.~\eqref{eq:loss:approx} as a column subset selection problem (CSSP) ~\cite{BoutsidisMD09, FarahatEGK15}. Accordingly, given the visual embedding matrix $\mathbf{X}_v \in \mathbb{R}^{N \times d}$, our aim is to select a subset of $k$ columns of $\mathbf{X}_v^\top \in \mathbb{R}^{d \times N}$ (\ie, tokens) whose span best approximates the original feature subspace. Using the Frobenius norm, the objective can be written as:
\begin{equation}
\label{eq3}
\min_{\mathcal{S} \subset \{1,\dots,N\}} \;
\left\| \mathbf{X}_v^\top - \mathbf{\Pi}_{\mathcal{S}} \mathbf{X}_v^\top \right\|_F^2
\quad \text{s.t. } |\mathcal{S}| = k.
\end{equation}
where $\mathbf{\Pi}_{\mathcal{S}} = 
\tilde{\mathbf{X}}_v^\top
(\tilde{\mathbf{X}}_v \tilde{\mathbf{X}}_v^\top)^{-1}
\tilde{\mathbf{X}}_v \in \mathbb{R}^{d \times d}$ 
denotes the orthogonal projection matrix onto the column space spanned by the selected visual tokens.

Since the CSSP is NP-hard~\cite{SHITOV202152}, exact optimization of Eq.~\eqref{eq3} is computationally infeasible.
We therefore employ a greedy subspace approximation method based on \textit{rank-revealing QR} (RRQR) factorization~\cite{chan1987rank}. RRQR iteratively selects tokens whose embeddings exhibit the largest projection residual with respect to the span of the currently selected subset.
To evaluate this residual, the approximation subspace must be explicitly represented, and hence we maintain an orthonormal basis for the selected tokens using the \textit{Gram--Schmidt process} \cite{BJORCK1994297}. 
The procedure proceeds as follows, as illustrated in the top-right panel of Fig.~\ref{fig2_overview}:
\begin{enumerate}
\item \textbf{Setup.}  
Let $\mathcal{S}$ and $\mathcal{Q}$ denote the selected token indices and the orthonormal basis, respectively.

\item \textbf{Compute residuals.}  
For each candidate token (\ie, column) $\mathbf{x}_i$, compute the residual $r_i$ with respect to the current subspace as:
\begin{equation}
\label{eq:residual}
r_i 
= 
\left\| \mathbf{x}_i - \mathbf{\Pi}_{\mathcal{S}} \mathbf{x}_i \right\|_2^2
=
\left\| \mathbf{x}_i \right\|_2^2
-
\sum_{\mathbf{q} \in \mathcal{Q}}
\left( \mathbf{x}_i^\top \mathbf{q} \right)^2.
\end{equation}

\item \textbf{Select token.}  
Choose the token with the largest residual and add its index to $\mathcal{S}$.

\item \textbf{Update basis.}  
Orthonormalize the selected token against the existing basis using Gram-Schmidt and append the resulting normalized vector to $\mathcal{Q}$.

\item \textbf{Iterate.}  
Repeat the above steps until $|\mathcal{S}| = k$.
\end{enumerate}
Through this iterative process, the span of the selected tokens progressively approximates the dominant subspace of the original visual embeddings.
Such RRQR-based greedy selection is known to admit provable approximation guarantees for the CSSP objective \cite{BoutsidisMD09} while remaining computationally efficient and deterministic.

\smalltitle{Theoretical Bound.}
The reconstruction objective in Eq.~\eqref{eq3} minimizes the perturbation introduced by token pruning at the visual embedding level. 
To relate this to the ideal objective in Eq.~\eqref{eq:loss:ideal}, which measures the discrepancy between LLM outputs, we adopt the standard Lipschitz stability perspective commonly used in neural network analysis~\cite{FazlyabRHMP19}. 
Specifically, assuming the LLM output function $\mathcal F(\cdot,\cdot)$ is locally $L$-Lipschitz with respect to perturbations in the visual embeddings, we obtain:
\begin{equation}
\left\|
\mathcal F(\tilde{\mathbf X}_v, \mathbf{X}_t)
-
\mathcal F(\mathbf X_v, \mathbf{X}_t)
\right\|
\le
L
\left\|
\tilde{\mathbf X}_v
-
\mathbf X_v
\right\|.
\end{equation}
This bound suggests that minimizing the reconstruction error of the visual embeddings helps control the change in the output distribution of the LLM.

\subsection{Anti-Relevance Guided Token Selection}
\label{sec:anti}

\begin{figure}[t]
    \centering
         \includegraphics[width=0.48\textwidth]{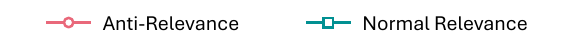} \\
    \includegraphics[width=0.99\linewidth]{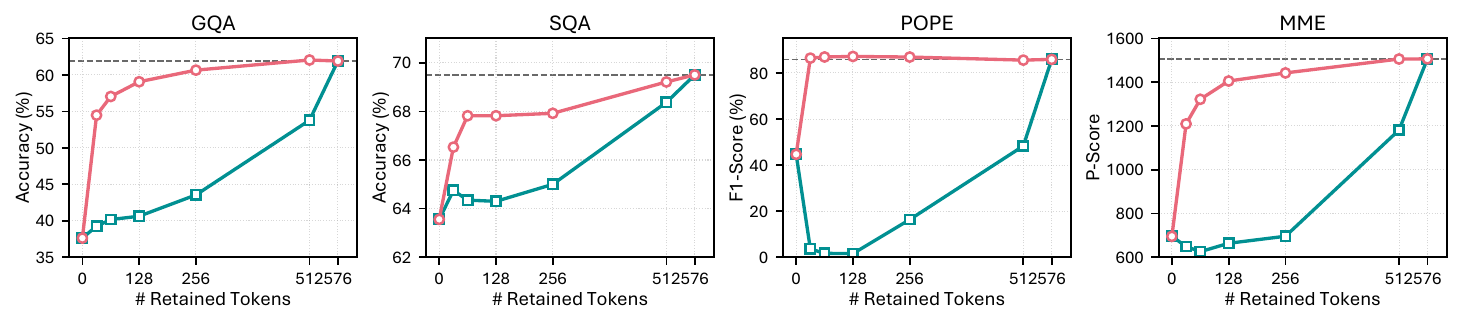}
        \vspace{-2mm}
    \caption{\textbf{Impact of Normal-Relevance and Anti-Relevance in Isolation.} Performance comparison of \textit{top-k} visual token selection based on image--text normal-relevance or anti-relevance across varying token retention levels on four benchmarks.}
    \label{fig3_impact_relevance}
\end{figure}

\begin{figure}[t]
    \centering
    \includegraphics[width=0.99\linewidth]{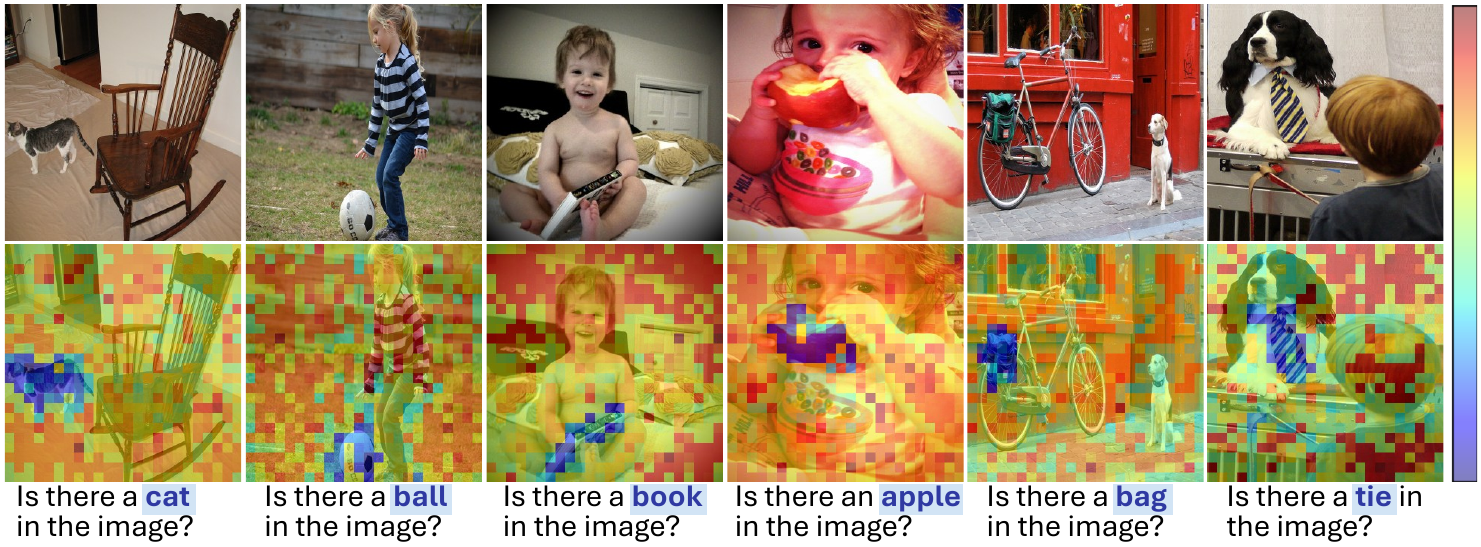}
        \vspace{-2mm}
    \caption{\textbf{Visualizations.} Visualization of image--text relevance scores on the POPE benchmark~\cite{LiDZWZW23}. Red indicates higher relevance, while blue indicates lower relevance.
    }
    \label{fig4_vis_relevance}
            \vspace{-2mm}
\end{figure}

\smalltitle{Anti-Relevance Phenomenon.}
Although our reconstruction-based strategy effectively captures the structure of the visual embedding space, it does not utilize any information from the textual query. This raises a crucial yet unresolved question: \textit{which visual tokens should be retained to best answer the query}.
To investigate this, we analyze how image–text relevance scores influence performance by varying both the token selection criterion and the number of retained tokens. The relevance score is measured as the cosine similarity between image and text token embeddings in the CLIP space~\cite{RadfordKHRGASAM21}, as commonly adopted in prior works~\cite{SongWCWGW25, zhang2025beyond}.
For clarity, we consider two extreme criteria without any additional guidance: \textit{normal-relevance} and \textit{anti-relevance}. The former preferentially selects tokens with higher relevance scores, whereas the latter prioritizes those with lower relevance scores. 
Contrary to the common intuition that \textit{normal-relevance} would outperform \textit{anti-relevance}, the empirical results in Fig.~\ref{fig3_impact_relevance} show the opposite trend. Specifically, \textit{anti-relevance} consistently outperforms \textit{normal-relevance} by a large margin across all settings. Moreover, the performance of \textit{normal-relevance} drops sharply even after pruning only a small number of visual tokens, even though these tokens are supposed to be more \textit{relevant} according to CLIP-based similarity scores.
This phenomenon is further supported by the visualization in Fig.~\ref{fig4_vis_relevance}, where semantically meaningful regions tend to receive relatively low relevance scores, while background areas unrelated to the query often exhibit higher relevance scores. This suggests that tokens highly aligned with the query often correspond to less informative regions in the CLIP space, whereas tokens with lower similarity may contain complementary contextual information that is crucial for reasoning.

\smalltitle{Anti-Relevance Guidance.}
Motivated by the empirical observation above, we introduce an anti-relevance score that explicitly favors visual tokens with lower similarity to the textual query. For each $n$-th visual token, the anti-relevance score is defined as:
\begin{equation*}
\rho_n = \frac{1}{M}\sum_{j=1}^{M} \big(-\cos(\mathbf{E}_{I,n}, \mathbf{E}_{T,j})\big), 
\quad \forall n \in \{1,\dots,N\},
\end{equation*}
where $\mathbf{E}_{I,n}$ and $\mathbf{E}_{T,j}$ denote the CLIP image and text embeddings, respectively, and $M$ is the number of text tokens. This formulation assigns larger values to tokens with lower image–text similarity, thereby emphasizing anti-relevance. Since absolute similarity values may vary significantly across images, we apply z-score normalization within each image to capture token-wise deviations from the image-specific similarity distribution. The normalized scores are then transformed using the softplus function to ensure non-negative weights compatible with the reconstruction-based selection process.

Finally, the anti-relevance scores are incorporated into the reconstruction-based token selection as a multiplicative guidance term. Specifically, at each iteration we select the token index that maximizes the product of the reconstruction residual $r_n$ and the anti-relevance score $\rho_n$:
\begin{equation*}
n^* = \arg\max_{n \notin \mathcal{S}} \left( r_n \rho_n \right),
\end{equation*}
where $n$ denotes a visual token index and $\mathcal{S}$ is the set of indices corresponding to already selected tokens. This formulation favors tokens that are both important for reconstructing the visual subspace and informative from the anti-relevance perspective.

\begin{wrapfigure}{R}{0.50\textwidth}
\vspace{-3.8\baselineskip}
\begin{minipage}{0.50\textwidth}

\begin{algorithm}[H]
\footnotesize
\caption{SPARE}
\label{alg:SPARE}

\textbf{Input}: Visual token matrix $\mathbf{X}_v \in \mathbb{R}^{N \times d}$,
image embeddings $\mathbf{E}_I \in \mathbb{R}^{N \times C}$,
text embeddings $\mathbf{E}_T \in \mathbb{R}^{M \times C}$, budget $k \le N$ \\
\textbf{Output}: Selection mask $\mathbf{S} \in \{0,1\}^{N}$

\begin{algorithmic}[1]
\State $\mathbf{S} \gets \mathbf{0}_N$; \ $\mathcal{Q} \gets \emptyset$

\State \textcolor{mygray}{\textit{------ Anti-Relevance Computation ------}}
\State $\rho_n \gets \frac{1}{M}\sum_{j=1}^{M}(-\cos(\mathbf{E}_{I,n},\mathbf{E}_{T,j})),\ \forall n$
\State $\rho_n \gets 
\mathrm{softplus}\!\left(
\frac{\rho_n-\mu_{\rho}}{\sigma_{\rho}}
\right),\ \forall n$
\State \textcolor{mygray}{\textit{--------- Visual Token Selection ---------}}
\State $r_n \gets \lVert \mathbf{X}_{v,n} \rVert_2^2,\ \forall n$

\For{$t = 1$ \textbf{to} $k$}
  \State $p \gets 
  \arg\max_{n \in \{ i \mid \mathbf{S}_i = 0 \}}
  \left( r_n \rho_n \right)$
  \State $\mathbf{S}_p \gets 1$; \ $\mathbf{v} \gets \mathbf{X}_{v,p}$
  \For{$\mathbf{q} \in \mathcal{Q}$}
    \State $\mathbf{v} \gets \mathbf{v} - (\mathbf{q} \cdot \mathbf{v})\,\mathbf{q}$
  \EndFor
  \State $\mathbf{q} \gets \mathbf{v} / \lVert \mathbf{v} \rVert_2$
  \State $\mathcal{Q} \gets \mathcal{Q} \cup \{\mathbf{q}\}$
  \State $r_n \gets r_n - (\mathbf{X}_{v,n} \cdot \mathbf{q})^2,\ \forall n \text{ s.t. } \mathbf{S}_n=0$
\EndFor

\State \Return $\mathbf{S}$
\end{algorithmic}

\end{algorithm}

\end{minipage}
\vspace{-1.8\baselineskip}
\end{wrapfigure}

\subsection{SPARE: Overall Process}

In summary, the overall pipeline of SPARE is illustrated in Fig.~\ref{fig2_overview}, and the detailed procedure is described in Algorithm~\ref{alg:SPARE}. SPARE starts by computing anti-relevance scores for all visual tokens using CLIP similarity (Lines 3--4), as described in Section~\ref{sec:anti}. Then, it performs visual token selection by jointly considering subspace reconstruction and anti-relevance. To reconstruct the original visual subspace, SPARE iteratively computes the residual of each token with respect to the span of the currently selected subset and selects the token with the largest residual. The initial residuals are set to their norms according to Eq.~\eqref{eq:residual} (Line 6).
Simultaneously, the anti-relevance scores are combined to guide the selection toward tokens that provide complementary contextual information beyond direct query alignment (Line 8). This joint criterion enables SPARE to preserve the structural representation of the visual embedding space while prioritizing tokens that provide complementary contextual information for query understanding.

\section{Experiments}

\begin{table}[t]
\caption{\textbf{Performance Comparison on LLaVA-1.5-7B.}}
\centering
\small
\setlength{\tabcolsep}{3.8pt}
\renewcommand{\arraystretch}{1.05}

\resizebox{\linewidth}{!}{%
\begin{tabular}{l|ccccccccc|c}
\toprule
\textbf{Method} &
\textbf{VQA$^{\mathrm{V2}}$} & \textbf{GQA} & \textbf{SQA$^{\mathrm{Img}}$} & \textbf{VQA$^{\mathrm{Text}}$} &
\textbf{POPE} & \textbf{MME} & \textbf{MMB$^{\mathrm{EN}}$} & \textbf{MMB$^{\mathrm{CN}}$} & \textbf{MMVet} & \textbf{Rel.\ Acc} \\

\rowcolor{gray!15}
\multicolumn{11}{c}{\textit{Total 576 tokens}}\\

LLaVA-1.5-7B
& 78.5 & 61.9 & 69.5 & 58.2 & 85.9 & 1506.5 & 64.7 & 58.1 & 31.3 & 100.0\% \\

\rowcolor{gray!15}
\multicolumn{11}{c}{\textit{Retain 128 tokens \textcolor{green!70!black}{$(\downarrow 77.8\%)$}}}\\

SparseVLM ({\color{black!60}ICML 25})
& 76.3 & 58.4 & 68.5 & \textbf{56.7} & 85.0 & 1428.9 & \textbf{64.3} & \textbf{58.2} & 28.9 & 97.0 \\
PruMerge+ ({\color{black!60}ICCV 25})
& 74.0 & 57.6 & 68.2 & 54.8 & 81.0 & 1376.9 & 60.7 & 54.6 & 27.7 & 93.5 \\
DART ({\color{black!60}EMNLP 25})
& 76.0 & 58.7 & \textbf{69.2} & 56.5 & 80.2 & \textbf{1476.3} & 63.1 & 57.3 & 28.7 & 96.4 \\
CDPruner ({\color{black!60}NeurIPS 25})
& 76.6 & 59.8 & 69.0 & 56.2 & \textbf{87.6} & 1437.8 & 63.0 & 55.2 & 32.0 & 98.0 \\
\textbf{SPARE (Ours)}
& \textbf{76.7} & \textbf{60.0} & 68.6 & 56.6 & 87.5 & 1415.6 & 62.5 & 56.4 & \textbf{32.8} & \textbf{\textcolor{red}{98.3}} \\

\rowcolor{gray!15}
\multicolumn{11}{c}{\textit{Retain 64 tokens \textcolor{green!70!black}{$(\downarrow 88.9\%)$}}}\\

SparseVLM ({\color{black!60}ICML 25})
& 70.3 & 53.7 & \textbf{69.7} & 53.5 & 77.6 & 1289.5 & 60.2 & 52.6 & 24.3 & 89.5 \\
PruMerge+ ({\color{black!60}ICCV 25})
& 71.5 & 55.2 & 68.0 & 53.7 & 75.8 & 1325.2 & 59.2 & 52.3 & 26.3 & 90.2 \\
DART ({\color{black!60}EMNLP 25})
& 72.8 & 56.3 & 68.8 & 54.3 & 74.2 & \textbf{1408.3} & \textbf{61.6} & 53.9 & 26.6 & 92.1 \\
CDPruner ({\color{black!60}NeurIPS 25})
& 75.4 & 58.6 & 68.1 & 55.2 & \textbf{87.5} & 1408.2 & 61.0 & 53.4 & 29.8 & 95.6 \\
\textbf{SPARE (Ours)}
& \textbf{75.8} & \textbf{59.1} & 68.1 & \textbf{55.8} & 87.3 & 1400.3 & 61.5 & \textbf{54.0} & \textbf{31.6} & \textbf{\textcolor{red}{96.6}} \\

\rowcolor{gray!15}
\multicolumn{11}{c}{\textit{Retain 32 tokens \textcolor{green!70!black}{$(\downarrow 94.4\%)$}}}\\

DART ({\color{black!60}EMNLP 25})
& 67.9 & 52.8 & 69.2 & 52.0 & 65.5 & 1298.0 & 58.8 & 48.8 & 22.7 & 85.6 \\
CDPruner ({\color{black!60}NeurIPS 25})
& 73.6 & 56.9 & \textbf{69.3} & 53.2 & \textbf{87.8} & \textbf{1377.8} & 59.5 & 49.6 & 27.9 & 93.0 \\
\textbf{SPARE (Ours)}
& \textbf{74.0} & \textbf{57.5} & \textbf{69.3} & \textbf{54.0} & 87.5 & 1364.3 & \textbf{60.6} & \textbf{50.3} & \textbf{31.4} & \textbf{\textcolor{red}{94.7}} \\
\bottomrule
\end{tabular}
}
\label{tab1_llava7b}
\end{table}

\begin{table}[t]
\caption{\textbf{Performance Comparison on LLaVA-1.5-13B.}}
\centering
\small
\setlength{\tabcolsep}{3.8pt}
\renewcommand{\arraystretch}{1.05}

\resizebox{\linewidth}{!}{%
\begin{tabular}{l|ccccccccc|c}
\toprule
\textbf{Method} &
\textbf{VQA$^{\mathrm{V2}}$} & \textbf{GQA} & \textbf{SQA$^{\mathrm{Img}}$} & \textbf{VQA$^{\mathrm{Text}}$} &
\textbf{POPE} & \textbf{MME} & \textbf{MMB$^{\mathrm{EN}}$} & \textbf{MMB$^{\mathrm{CN}}$} & \textbf{MMVet} & \textbf{Rel.\ Acc} \\

\rowcolor{gray!15}
\multicolumn{11}{c}{\textit{Total 576 tokens}}\\
LLaVA-1.5-13B
& 80.0 & 63.3 & 72.8 & 61.2 & 86.0 & 1531.2 & 68.5 & 63.5 & 36.2 & 100.0\% \\

\rowcolor{gray!15}
\multicolumn{11}{c}{\textit{Retain 128 tokens \textcolor{green!70!black}{$(\downarrow 77.8\%)$}}}\\

SparseVLM ({\color{black!60}ICML 25})
& 77.1 & 58.7 & 74.1 & \textbf{59.1} & 84.2 & 1494.3 & \textbf{68.5} & 63.1 & 35.7 & 97.9 \\
PruMerge+ ({\color{black!60}ICCV 25})
& 75.1 & 57.4 & 71.8 & 56.6 & 80.9 & 1403.2 & 64.9 & 60.2 & 31.8 & 93.2 \\
DART ({\color{black!60}EMNLP 25})
& 77.6 & \textbf{61.0} & \textbf{74.4} & 59.0 & 81.9 & \textbf{1534.8} & 67.6 & \textbf{63.3} & 31.9 & 97.1 \\
CDPruner ({\color{black!60}NeurIPS 25})
& 77.7 & 59.8 & 72.6 & 58.4 & 86.9 & 1468.5 & 67.5 & 61.4 & 37.1 & 97.9 \\
\textbf{SPARE (Ours)}
& \textbf{78.0} & 59.7 & 72.8 & 58.6 & \textbf{87.3} & 1449.8 & 66.9 & 62.1 & \textbf{37.3} & \textbf{\textcolor{red}{98.0}} \\

\rowcolor{gray!15}
\multicolumn{11}{c}{\textit{Retain 64 tokens \textcolor{green!70!black}{$(\downarrow 88.9\%)$}}}\\

SparseVLM ({\color{black!60}ICML 25})
& 71.3 & 55.7 & 72.5 & 55.8 & 76.2 & 1353.5 & 63.0 & 58.6 & 28.4 & 89.7 \\
PruMerge+ ({\color{black!60}ICCV 25})
& 72.4 & 55.6 & 71.7 & 55.7 & 74.1 & 1332.5 & 63.9 & 58.9 & 30.1 & 90.0 \\
DART ({\color{black!60}EMNLP 25})
& 73.7 & 57.1 & \textbf{73.9} & 55.8 & 75.6 & 1433.0 & 65.5 & \textbf{61.0} & 30.4 & 92.5 \\
CDPruner ({\color{black!60}NeurIPS 25})
& 76.7 & \textbf{59.4} & 72.5 & 57.6 & \textbf{87.1} & 1436.7 & 65.5 & 58.9 & 36.0 & 96.3 \\
\textbf{SPARE (Ours)}
& \textbf{77.1} & \textbf{59.4} & 72.3 & \textbf{58.1} & 87.0 & \textbf{1467.4} & \textbf{66.2} & 60.1 & \textbf{37.0} & \textbf{\textcolor{red}{97.2}} \\

\rowcolor{gray!15}
\multicolumn{11}{c}{\textit{Retain 32 tokens \textcolor{green!70!black}{$(\downarrow 94.4\%)$}}}\\

DART ({\color{black!60}EMNLP 25})
& 68.3 & 54.0 & 72.8 & 52.7 & 66.7 & 1319.7 & 61.9 & \textbf{57.0} & 26.5 & 86.0 \\
CDPruner ({\color{black!60}NeurIPS 25})
& 75.2 & \textbf{58.4} & 71.9 & 55.2 & 87.6 & 1417.3 & \textbf{63.7} & 56.6 & 29.5 & 92.6 \\
\textbf{SPARE (Ours)}
& \textbf{75.7} & \textbf{58.4} & \textbf{73.2} & \textbf{56.3} & \textbf{87.7} & \textbf{1420.9} & \textbf{63.7} & 56.5 & \textbf{34.7} & \textbf{\textcolor{red}{94.7}} \\
\bottomrule
\end{tabular}
}
\label{tab2_llava13b}
\end{table}

We evaluate SPARE on diverse VLMs and benchmarks to examine three aspects: 
(i) its robustness under aggressive token reduction, 
(ii) its effectiveness for multi-skill compositional reasoning across different VLM architectures, and 
(iii) the contributions of subspace reconstruction and anti-relevance guidance.


\subsection{Experimental Setup}

\smalltitle{Models.}
We evaluate SPARE on several representative VLM backbones, including LLaVA-1.5-7B and LLaVA-1.5-13B~\cite{LiuLLL24} for standard image understanding, as well as LLaVA-NeXT-7B~\cite{liu2024llavanext}, which supports high-resolution inputs with up to 2,880 visual tokens per image. 
We further evaluate Qwen2.5-VL-7B~\cite{abs-2502-13923}, which adopts dynamic image resolutions and produces input-dependent numbers of visual tokens.

\smalltitle{Benchmarks.}
We evaluate on nine widely used benchmarks: VQAv2~\cite{GoyalKSBP17}, GQA~\cite{HudsonM19}, SQA-IMG~\cite{LuMX0CZTCK22}, TextVQA~\cite{SinghNSJCBPR19}, POPE~\cite{LiDZWZW23}, MME~\cite{fu2025mme}, MMBench-EN~\cite{LiuDZLZZYWHLCL24}, MMBench-CN~\cite{LiuDZLZZYWHLCL24}, and MM-Vet~\cite{YuYLWL0WW24}. 
These benchmarks cover diverse multimodal tasks, including visual question answering, OCR-based reasoning, hallucination evaluation, and multi-skill compositional reasoning.

\smalltitle{Implementation Details.}
All experiments are conducted on NVIDIA A6000 GPUs. Unless otherwise specified, we follow the default inference configurations of each backbone and apply SPARE in a training-free, plug-and-play manner. Additional implementation details are provided in the Appendix.




\subsection{Overall Results}

\smalltitle{LLaVA-1.5-7B and LLaVA-1.5-13B.}
As shown in Tables~\ref{tab1_llava7b} and~\ref{tab2_llava13b}, SPARE consistently outperforms state-of-the-art pruning methods across diverse benchmarks on both LLaVA-1.5-7B and LLaVA-1.5-13B~\cite{LiuLLL24}. The advantage becomes more pronounced under aggressive pruning settings (\ie, 94.4\% token reduction), where preserving the original visual representation becomes increasingly difficult. Even under such extreme reduction, SPARE retains about 95\% of the baseline performance and surpasses nearly all methods evaluated at 88.9\% pruning, except CDPruner~\cite{zhang2025beyond}. While CDPruner remains the strongest competitor, SPARE maintains a clear margin on several benchmarks, particularly on MM-Vet~\cite{YuYLWL0WW24}, which evaluates multi-skill compositional reasoning.


\smalltitle{LLaVA-NeXT-7B.}
We further evaluate SPARE under higher-resolution settings using LLaVA-NeXT-7B~\cite{liu2024llavanext}. As shown in Table~\ref{tab3_llava_next}, SPARE achieves strong overall performance across pruning ratios. When retaining 22.2\% of tokens (approximately 640 tokens), the improvement becomes less pronounced, likely because the large number of preserved tokens reduces the impact of token selection in this high-resolution setting. Nevertheless, under extreme pruning (\ie, 94.4\% token reduction), SPARE again demonstrates strong robustness, preserving about 96\% of the original performance and showing clear advantages on multi-skill compositional reasoning (\ie, MM-Vet~\cite{YuYLWL0WW24}).


\begin{table}[t]
\caption{\textbf{Performance Comparison on LLaVA-NeXT-7B.}}
\centering
\small
\setlength{\tabcolsep}{3.8pt}
\renewcommand{\arraystretch}{1.05}

\resizebox{\linewidth}{!}{%
\begin{tabular}{l|ccccccccc|c}
\toprule
\textbf{Method} &
\textbf{VQA$^{\mathrm{V2}}$} & \textbf{GQA} & \textbf{SQA$^{\mathrm{Img}}$} & \textbf{VQA$^{\mathrm{Text}}$} &
\textbf{POPE} & \textbf{MME} & \textbf{MMB$^{\mathrm{EN}}$} & \textbf{MMB$^{\mathrm{CN}}$} & \textbf{MMVet} & \textbf{Rel.\ Acc} \\

\rowcolor{gray!15}
\multicolumn{11}{c}{\textit{Retain 2880 tokens (100\%)}}\\
LLaVA-NeXT-7B
& 81.3 & 62.5 & 67.5 & 60.3 & 86.8 & 1511.8 & 65.8 & 57.3 & 40.0 & 100.0\% \\

\rowcolor{gray!15}
\multicolumn{11}{c}{\textit{Retain 640 tokens \textcolor{green!70!black}{$(\downarrow 77.8\%)$}}}\\
DART ({\color{black!60}EMNLP 25})
& \textbf{80.6} & \textbf{63.1} & \textbf{69.2} & \textbf{62.5} & 85.9 & \textbf{1486.4} & \textbf{66.5} & \textbf{59.4} & 38.1 & \textbf{\textcolor{red}{100.4}} \\
CDPruner ({\color{black!60}NeurIPS 25})
& 79.8 & 62.6 & 67.9 & 57.4 & 87.3 & 1461.5 & 66.2 & 57.6 & \textbf{40.1} & 99.2 \\
\textbf{SPARE (Ours)}
& 79.9 & 62.6 & 68.0 & 59.0 & \textbf{87.4} & 1480.4 & \textbf{66.5} & 57.7 & 37.7 & 99.1 \\

\rowcolor{gray!15}
\multicolumn{11}{c}{\textit{Retain 320 tokens \textcolor{green!70!black}{$(\downarrow 88.9\%)$}}}\\
DART ({\color{black!60}EMNLP 25})
& \textbf{78.5} & \textbf{61.4} & \textbf{68.3} & \textbf{58.3} & 83.5 & 1421.6 & \textbf{65.5} & 56.7 & 37.1 & 97.1 \\
CDPruner ({\color{black!60}NeurIPS 25})
& 78.4 & \textbf{61.4} & 67.6 & 57.4 & \textbf{87.3} & \textbf{1471.1} & \textbf{65.5} & 55.5 & 37.6 & 97.6 \\
\textbf{SPARE (Ours)}
& 78.4 & 61.3 & 67.6 & 57.5 & 87.2 & 1451.5 & 65.3 & \textbf{56.9} & \textbf{39.6} & \textbf{\textcolor{red}{98.2}} \\

\rowcolor{gray!15}
\multicolumn{11}{c}{\textit{Retain 160 tokens \textcolor{green!70!black}{$(\downarrow 94.4\%)$}}}\\
DART ({\color{black!60}EMNLP 25})
& 73.7 & 57.3 & \textbf{68.4} & 50.1 & 77.1 & 1388.4 & 60.9 & 52.9 & 31.4 & 90.1 \\
CDPruner ({\color{black!60}NeurIPS 25})
& \textbf{76.7} & 60.8 & 67.1 & 55.3 & \textbf{86.9} & \textbf{1436.2} & 64.2 & 54.4 & 35.2 & 95.4 \\
\textbf{SPARE (Ours)}
& 76.6 & \textbf{61.0} & 67.5 & \textbf{55.9} & 86.4 & 1409.9 & \textbf{64.3} & \textbf{55.2} & \textbf{37.7} & \textbf{\textcolor{red}{96.2}} \\

\bottomrule
\end{tabular}
}
\label{tab3_llava_next}
\end{table}

\begin{table}[t]
\centering
\small

\begin{minipage}[t]{0.48\linewidth}
\caption{\textbf{Performance Comparison on Multi-Skill Benchmark.} Grouped by the number of required reasoning skills (1–4), reporting the average performance for each corresponding skill count.}
\label{tab4_multi_skill}
\centering
\setlength{\tabcolsep}{6pt}
\renewcommand{\arraystretch}{1.08}

\resizebox{\linewidth}{!}{%
\begin{tabular}{l|cccc}
\toprule
\textbf{Method} &
\textbf{1-Skill} &
\textbf{2-Skill} &
\textbf{3-Skill} &
\textbf{4-Skill} \\
\rowcolor{gray!15}
\multicolumn{5}{c}{\textit{Retain 32 tokens}} \\
CDPruner ({\color{black!60}NeurIPS 25})
& \textbf{50.4} & 27.8 & 6.5 & 17.9 \\
\textbf{SPARE (Ours)}
& 44.7 & \textbf{32.3} & \textbf{21.1} & \textbf{22.9} \\

\bottomrule
\end{tabular}
}
    \vspace{-4mm}
\end{minipage}
\hfill
\begin{minipage}[t]{0.48\linewidth}
\caption{\textbf{Performance Comparison on Qwen2.5-VL-7B.}}
\vspace{-2mm}
\label{tab5_qwen}
\centering
\setlength{\tabcolsep}{6pt}
\renewcommand{\arraystretch}{1.08}

\resizebox{\linewidth}{!}{%
\begin{tabular}{l|cccc}
\toprule
\textbf{Method} &
\textbf{GQA} &
\textbf{SQA} &
\textbf{POPE} &
\textbf{MME} \\
\rowcolor{gray!15}
\multicolumn{5}{c}{\textit{All tokens (100\%)}} \\
Qwen2.5-VL-7B
& 60.9 & 88.4 & 86.3 & 2307.6 \\
\rowcolor{gray!15}
\multicolumn{5}{c}{\textit{Retain 20\% tokens}} \\
CDPruner ({\color{black!60}NeurIPS 25})
& 58.7 & 84.3 & 84.3 & \textbf{2166.9} \\
\textbf{SPARE (Ours)}
& \textbf{59.4} & \textbf{84.9} & \textbf{85.3} & 1997.3 \\
\rowcolor{gray!15}
\multicolumn{5}{c}{\textit{Retain 10\% tokens}} \\
CDPruner ({\color{black!60}NeurIPS 25})
& 56.0 & 82.1 & 80.0 & \textbf{2008.7} \\
\textbf{SPARE (Ours)}
& \textbf{57.2} & \textbf{82.5} & \textbf{83.7} & 1923.3 \\
\bottomrule
\end{tabular}
}
    \vspace{-4mm}
\end{minipage}

\end{table}

\begin{table}[t]
\centering
\small

\begin{minipage}[t]{0.48\linewidth}
\caption{\textbf{Reconstruction Error.} Measured as the Frobenius norm between the feature matrices before and after pruning.}
\label{tab6_frobenius_error}
\centering
\setlength{\tabcolsep}{6pt}
\renewcommand{\arraystretch}{1.08}

\resizebox{\linewidth}{!}{%
\begin{tabular}{l|cccc}
\toprule
\textbf{Method} & 
\textbf{GQA$\downarrow$} & 
\textbf{SQA$\downarrow$} & 
\textbf{POPE$\downarrow$} & 
\textbf{MME$\downarrow$} \\

\rowcolor{gray!15}
\multicolumn{5}{c}{\textit{Reconstruction Error (Retain 32 tokens)}}\\

DART ({\color{black!60}EMNLP 25})
& 398.4 & 376.2 & 417.9 & 409.3 \\

CDPruner ({\color{black!60}NeurIPS 25})
& 240.8 & 210.4 & 239.1 & 229.5 \\

\textbf{SPARE (Ours)}
& \textbf{228.2} & \textbf{178.9} & \textbf{224.7} & \textbf{209.5} \\

\bottomrule
\end{tabular}
}
    \vspace{-4mm}
\end{minipage}
\hfill
\begin{minipage}[t]{0.48\linewidth}
\caption{\textbf{Efficiency Comparison with Different Pruning Methods on LLaVA-1.5-7B.}}
\label{tab7_efficiency}
\centering
\setlength{\tabcolsep}{4pt}
\renewcommand{\arraystretch}{1.05}
\vspace{-1mm}

\resizebox{\linewidth}{!}{%
\begin{tabular}{l|cccc}
\toprule
\multirow{2}{*}{\textbf{Method}} &
\multirow{2}{*}{\textbf{\# Tokens}} &
\textbf{Prefill} &
\textbf{KV Cache} &
\textbf{Acc} \\
& & \textbf{(ms/sample)} & \textbf{(MB)} & \textbf{(\%)} \\
\midrule

\rowcolor{gray!15}
LLaVA-1.5-7B
& 576 & 130 & 347.7 & 58.2 \\

DART ({\color{black!60}EMNLP 25})
& \textbf{64} & 60 & 109.6 & 54.3 \\

CDPruner ({\color{black!60}NeurIPS 25})
& \textbf{64} & \textbf{45} & \textbf{91.8} & 55.2 \\

\textbf{SPARE (Ours)}
& \textbf{64} & \textbf{45} & \textbf{91.8} & \textbf{55.8} \\

\bottomrule
\end{tabular}
}
    \vspace{-4mm}
\end{minipage}
\label{tab:combined_tables}

\end{table}

\smalltitle{Multi-Skill Compositional Tasks.}
As observed in Tables \ref{tab1_llava7b}--\ref{tab3_llava_next}, SPARE shows a particularly large advantage on MM-Vet \cite{YuYLWL0WW24}, a multi-skill reasoning benchmark. To better understand this gap, we conduct a more detailed analysis by grouping questions according to the number of required reasoning skills. Using the LLaVA-1.5-7B results with 32 retained visual tokens, we compare the average performance of SPARE and CDPruner~\cite{zhang2025beyond} across these groups. As shown in Table~\ref{tab4_multi_skill}, CDPruner performs better than SPARE only on single-skill questions (e.g., recognition or OCR). However, as the number of required skills increases, SPARE shows increasingly larger gains. This trend highlights the advantage of our reconstruction-based token selection in preserving complementary contextual information necessary for compositional reasoning. See the Appendix for the full results.


\smalltitle{Qwen2.5-VL-7B.}
To further evaluate SPARE on a recent VLM, we conduct experiments on Qwen2.5-VL-7B~\cite{abs-2502-13923}. Unlike LLaVA, Qwen adopts dynamic-resolution visual processing, where the number of visual tokens varies across images. This property makes it difficult to directly apply CLIP-based image–text relevance, as CLIP~\cite{RadfordKHRGASAM21} operates on fixed-resolution inputs that are not naturally compatible with Qwen’s visual representations. Since CDPruner~\cite{zhang2025beyond} also relies on CLIP similarity, we remove image–text relevance from both methods to ensure a fair comparison. Under this setting, CDPruner reduces to a purely diversity-based selection strategy, allowing us to directly compare diversity-based selection with our reconstruction-based approach. As shown in Table~\ref{tab5_qwen}, the reconstruction-based strategy outperforms diversity-based selection on most benchmarks. These results suggest that the strength of SPARE stems not only from anti-relevance guidance but also from its principled reconstruction of the visual subspace, supporting the effectiveness of reconstruction-based token selection.


\subsection{Further Analysis}

\smalltitle{Reconstruction Error.}
We evaluate how well the selected tokens preserve the original visual representation by measuring the reconstruction error after pruning. Specifically, on LLaVA-1.5-7B~\cite{LiuLLL24}, we retain only 32 visual tokens and compute the Frobenius norm between the original and pruned feature matrices, averaged over images within each benchmark. To focus on the intrinsic reconstruction capability of each selection strategy, we consider a text-agnostic setting where image--text relevance is not used. 
Under this setting, we compare SPARE with diversity-based methods, including DART (diversity-only)~\cite{WenGWZZLHZ25} and CDPruner~\cite{zhang2025beyond}. As shown in Table~\ref{tab6_frobenius_error}, reconstruction-based selection consistently yields smaller reconstruction errors than diversity-based methods across multiple benchmarks. 
These results indicate that SPARE more faithfully preserves the original visual feature space, supporting the effectiveness of reconstruction-based token selection.



\begin{figure}[t]
    \centering
     \includegraphics[width=0.75\textwidth]{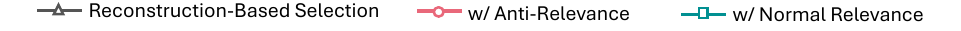} \\
    \centering
    \includegraphics[width=0.99\linewidth]{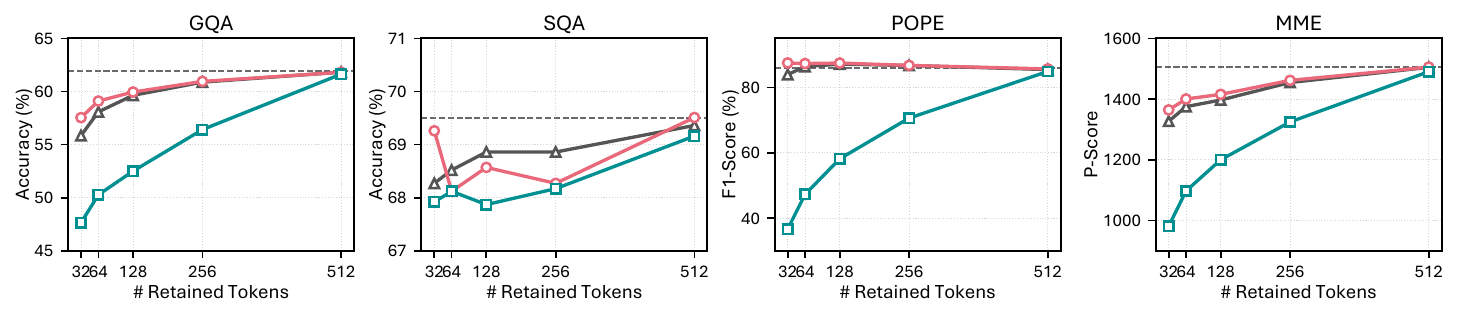}
        \vspace{-1mm}
    \caption{\textbf{Ablation Study of SPARE Components.} Comparison on LLaVA-1.5-7B \cite{LiuLLL24} across four benchmarks under three variants: (1) reconstruction-based selection, (2) with anti-relevance guidance, and (3) with normal-relevance guidance.}
    \label{fig5_ablation}
    \vspace{-2mm}
\end{figure}

\smalltitle{Efficiency Analysis.}
We now evaluate the efficiency of SPARE by measuring prefilling time and KV cache memory consumption. All experiments are conducted using LLaVA-1.5-7B while retaining 64 visual tokens on the TextVQA benchmark~\cite{SinghNSJCBPR19}. For reliable measurement, we perform a warm-up over 20 samples before evaluation, and report KV cache memory based on the peak usage during inference. As shown in Table~\ref{tab7_efficiency}, SPARE reduces the per-sample prefilling time by $2.9\times$ and decreases KV cache memory usage by $3.8\times$. Compared to other pruning methods, SPARE achieves the highest computational efficiency while maintaining competitive accuracy.


\smalltitle{Ablation Study.}
We conduct an ablation study to analyze the contributions of reconstruction-based selection and anti-relevance guidance in SPARE. Specifically, we compare three variants: (1) reconstruction-based selection alone, (2) reconstruction with anti-relevance guidance, and (3) reconstruction with normal-relevance guidance. All experiments are performed on LLaVA-1.5-7B~\cite{LiuLLL24}, evaluating accuracy under different visual token budgets.


As shown in Fig.~\ref{fig5_ablation}, reconstruction-based selection alone already exhibits strong robustness under aggressive pruning. On SQA and POPE, it retains only 32 visual tokens (\ie 94\% reduction) while preserving about 98\% of the baseline performance. Incorporating anti-relevance guidance further improves performance across most token budgets. In contrast, normal-relevance guidance consistently underperforms anti-relevance, supporting our design choice. Interestingly, on SQA~\cite{LuMX0CZTCK22}, relevance guidance does not always improve over reconstruction-only selection. We attribute this to the nature of SQA questions, which often require holistic understanding of scientific diagrams rather than focusing on text-aligned regions. This observation highlights the potential need for task-adaptive token selection strategies, which we leave for future work.


\begin{figure}[t]
    \centering
    \includegraphics[width=0.95\linewidth]{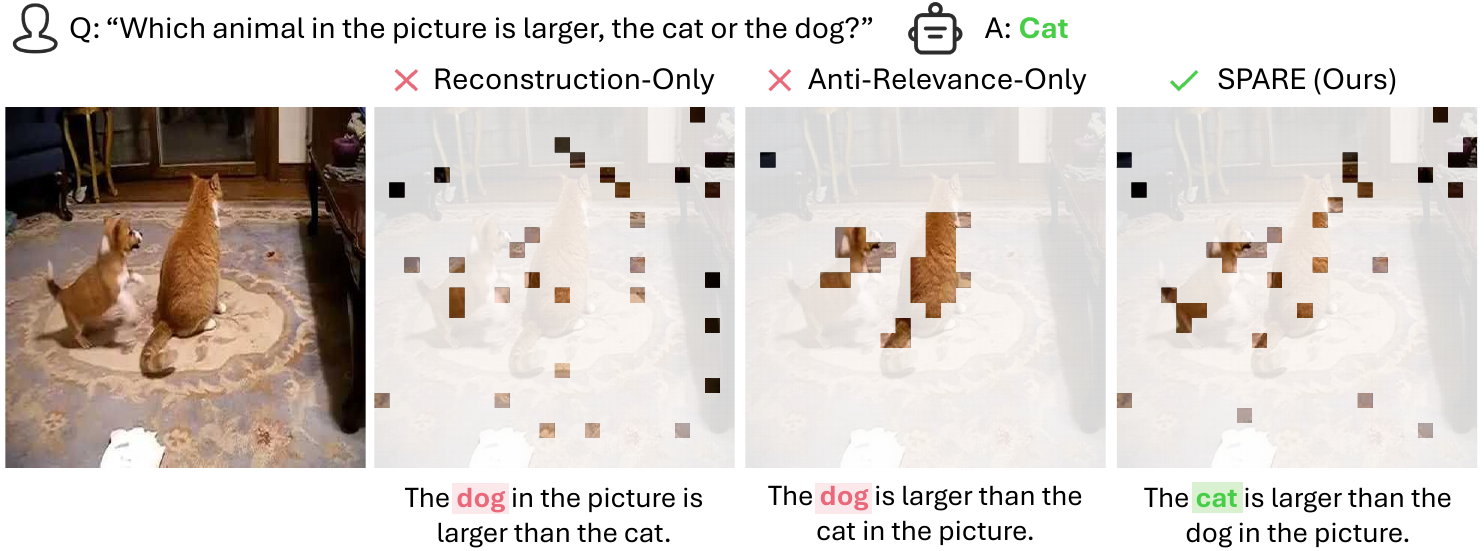}
        \vspace{-1mm}
    \caption{\textbf{Visualizations of SPARE.} Reconstruction-based selection focuses on structure, Anti-Relevance captures context but is redundant, while SPARE balances both.}
    \label{fig6_vis}
    \vspace{-2mm}
\end{figure}

\smalltitle{Visualizations.}
Lastly, we provide qualitative visualizations to better understand the roles of SPARE’s components. 
Specifically, we visualize the tokens retained by reconstruction-based selection alone and anti-relevance alone (via top-$k$ selection), while keeping 32 tokens on LLaVA-1.5-7B~\cite{LiuLLL24}. As shown in Fig.~\ref{fig6_vis}, reconstruction-based selection primarily preserves the global visual structure but may overlook regions closely aligned with the textual query. 
In contrast, anti-relevance captures query-related regions but often produces redundant selections and may miss visually important areas that are not directly aligned with the text (e.g., object boundaries required for size comparison). 
SPARE combines the strengths of both components, preserving the visual structure while selecting contextually informative tokens, which leads to strong question-answering performance. 
Additional visualizations are provided in the Appendix.

\section{Conclusion}

Visual token pruning is often treated as a heuristic selection problem. In this work, we revisit it from a structural perspective and formulate visual token pruning as a column subset selection problem. Based on this formulation, we propose SPARE, a token pruning method that selects tokens by reconstructing the original visual feature subspace through a rank-revealing QR procedure. Beyond structural reconstruction, we identify an anti-relevance phenomenon: tokens with lower image–text relevance often contain complementary contextual cues that are critical for reasoning. By integrating this anti-relevance guidance with reconstruction-based selection, SPARE balances global structural preservation and query-aware token prioritization. Extensive experiments demonstrate that SPARE remains robust even under extreme token reduction (\eg, 94\%) and consistently improves performance on challenging multi-skill compositional reasoning tasks, while operating in a fully training-free manner. These findings suggest that preserving the visual subspace structure provides a principled foundation for efficient and reliable token selection in VLMs.


%
%
\bibliographystyle{splncs04}
\bibliography{main}

@inproceedings{AlvarSAZ25,
  author       = {Saeed Ranjbar Alvar and
                  Gursimran Singh and
                  Mohammad Akbari and
                  Yong Zhang},
  title        = {{DivPrune: Diversity-based Visual Token Pruning for Large Multimodal Models}},
  booktitle    = {{IEEE/CVF} Conference on Computer Vision and Pattern Recognition, {CVPR}},
  year         = {2025},
}

@article{BJORCK1994297,
title = {Numerics of Gram-Schmidt orthogonalization},
journal = {Linear Algebra and its Applications},
year = {1994},
author = {A. Bjorck},
}

@inproceedings{BoutsidisMD09,
  author       = {Christos Boutsidis and
                  Michael W. Mahoney and
                  Petros Drineas},
  title        = {{An Improved Approximation Algorithm for the Column Subset Selection Problem}},
  booktitle    = {Proceedings of the Twentieth Annual {ACM-SIAM} Symposium on Discrete Algorithms, {SODA}},
  year         = {2009},
}

@article{chan1987rank,
  title={{Rank Revealing QR Factorizations}},
  author={Chan, Tony F},
  journal={Linear algebra and its applications},
  year={1987},
  publisher={Elsevier}
}

@inproceedings{ChenZLBLZC24,
  author       = {Liang Chen and
                  Haozhe Zhao and
                  Tianyu Liu and
                  Shuai Bai and
                  Junyang Lin and
                  Chang Zhou and
                  Baobao Chang},
  title        = {{An Image is Worth 1/2 Tokens After Layer 2: Plug-and-Play Inference Acceleration for Large Vision-Language Models}},
  booktitle    = {18th European Conference on Computer Vision, {ECCV}},
  year         = {2024},
}

@inproceedings{DaoFERR22,
  author       = {Tri Dao and
                  Daniel Y. Fu and
                  Stefano Ermon and
                  Atri Rudra and
                  Christopher R{\'{e}}},
  title        = {{FlashAttention: Fast and Memory-Efficient Exact Attention with IO-Awareness}},
  booktitle    = {Advances in Neural Information Processing Systems 35: Annual Conference on Neural Information Processing Systems, {NeurIPS}},
  year         = {2022},
}

@article{DeshpandeRVW06,
  author       = {Amit Deshpande and
                  Luis Rademacher and
                  Santosh S. Vempala and
                  Grant Wang},
  title        = {{Matrix Approximation and Projective Clustering via Volume Sampling}},
  journal      = {Theory Comput.},
  year         = {2006},
}

@article{DrineasMM08,
  author       = {Petros Drineas and
                  Michael W. Mahoney and
                  S. Muthukrishnan},
  title        = {{Relative-Error {CUR} Matrix Decompositions}},
  journal      = {{SIAM} J. Matrix Anal. Appl.},
  year         = {2008},
}

@article{FarahatEGK15,
  author       = {Ahmed K. Farahat and
                  Ahmed Elgohary and
                  Ali Ghodsi and
                  Mohamed S. Kamel},
  title        = {{Greedy Column Subset Selection for Large-Scale Data Sets}},
  journal      = {Knowl. Inf. Syst.},
  year         = {2015},
}

@inproceedings{fu2025mme,
title={{MME}: A Comprehensive Evaluation Benchmark for Multimodal Large Language Models},
author={Chaoyou Fu and Peixian Chen and Yunhang Shen and Yulei Qin and Mengdan Zhang and Xu Lin and Jinrui Yang and Xiawu Zheng and Ke Li and Xing Sun and Yunsheng Wu and Rongrong Ji and Caifeng Shan and Ran He},
booktitle={The Thirty-ninth Annual Conference on Neural Information Processing Systems Datasets and Benchmarks Track},
year={2025},
}

@inproceedings{GoyalKSBP17,
  author       = {Yash Goyal and
                  Tejas Khot and
                  Douglas Summers{-}Stay and
                  Dhruv Batra and
                  Devi Parikh},
  title        = {Making the {V} in {VQA} Matter: Elevating the Role of Image Understanding
                  in Visual Question Answering},
  booktitle    = {2017 {IEEE} Conference on Computer Vision and Pattern Recognition,
                  {CVPR} 2017, Honolulu, HI, USA, July 21-26, 2017},
  year         = {2017},
}

@article{GuE96,
  author       = {Ming Gu and
                  Stanley C. Eisenstat},
  title        = {{Efficient Algorithms for Computing a Strong Rank-Revealing {QR} Factorization}},
  journal      = {{SIAM} J. Sci. Comput.},
  year         = {1996},
}

@inproceedings{HudsonM19,
  author       = {Drew A. Hudson and
                  Christopher D. Manning},
  title        = {{GQA:} {A} New Dataset for Real-World Visual Reasoning and Compositional
                  Question Answering},
  booktitle    = {{IEEE} Conference on Computer Vision and Pattern Recognition, {CVPR}
                  2019, Long Beach, CA, USA, June 16-20, 2019},
  year         = {2019},
}

@inproceedings{LiDZWZW23,
  author       = {Yifan Li and
                  Yifan Du and
                  Kun Zhou and
                  Jinpeng Wang and
                  Wayne Xin Zhao and
                  Ji{-}Rong Wen},
  title        = {{Evaluating Object Hallucination in Large Vision-Language Models}},
  booktitle    = {Proceedings of the 2023 Conference on Empirical Methods in Natural Language Processing, {EMNLP}},
  year         = {2023},
}

@inproceedings{LiuDZLZZYWHLCL24,
  author       = {Yuan Liu and
                  Haodong Duan and
                  Yuanhan Zhang and
                  Bo Li and
                  Songyang Zhang and
                  Wangbo Zhao and
                  Yike Yuan and
                  Jiaqi Wang and
                  Conghui He and
                  Ziwei Liu and
                  Kai Chen and
                  Dahua Lin},
  title        = {MMBench: Is Your Multi-modal Model an All-Around Player?},
  booktitle    = {Computer Vision - {ECCV} 2024 - 18th European Conference, Milan, Italy,
                  September 29-October 4, 2024, Proceedings, Part {VI}},
  year         = {2024},
}

@inproceedings{LiuLLL24,
  author       = {Haotian Liu and
                  Chunyuan Li and
                  Yuheng Li and
                  Yong Jae Lee},
  title        = {Improved Baselines with Visual Instruction Tuning},
  booktitle    = {{IEEE/CVF} Conference on Computer Vision and Pattern Recognition,
                  {CVPR} 2024, Seattle, WA, USA, June 16-22, 2024},
  year         = {2024},
}

@inproceedings{LiuLWL23a,
  author       = {Haotian Liu and
                  Chunyuan Li and
                  Qingyang Wu and
                  Yong Jae Lee},
  title        = {{Visual Instruction Tuning}},
  booktitle    = {Advances in Neural Information Processing Systems 36: Annual Conference on Neural Information Processing Systems, {NeurIPS}},
  year         = {2023},
}

@misc{liu2024llavanext,
    title={LLaVA-NeXT: Improved reasoning, OCR, and world knowledge},
    url={https://llava-vl.github.io/blog/2024-01-30-llava-next/},
    author={Liu, Haotian and Li, Chunyuan and Li, Yuheng and Li, Bo and Zhang, Yuanhan and Shen, Sheng and Lee, Yong Jae},
    year={2024}
}

@inproceedings{LuMX0CZTCK22,
  author       = {Pan Lu and
                  Swaroop Mishra and
                  Tanglin Xia and
                  Liang Qiu and
                  Kai{-}Wei Chang and
                  Song{-}Chun Zhu and
                  Oyvind Tafjord and
                  Peter Clark and
                  Ashwin Kalyan},
  title        = {Learn to Explain: Multimodal Reasoning via Thought Chains for Science
                  Question Answering},
  booktitle    = {Advances in Neural Information Processing Systems 35: Annual Conference
                  on Neural Information Processing Systems 2022, NeurIPS 2022, New Orleans,
                  LA, USA, November 28 - December 9, 2022},
  year         = {2022},
}

@inproceedings{RadfordKHRGASAM21,
  author       = {Alec Radford and
                  Jong Wook Kim and
                  Chris Hallacy and
                  Aditya Ramesh and
                  Gabriel Goh and
                  Sandhini Agarwal and
                  Girish Sastry and
                  Amanda Askell and
                  Pamela Mishkin and
                  Jack Clark and
                  Gretchen Krueger and
                  Ilya Sutskever},
  title        = {Learning Transferable Visual Models From Natural Language Supervision},
  booktitle    = {Proceedings of the 38th International Conference on Machine Learning,
                  {ICML} 2021, 18-24 July 2021, Virtual Event},
  year         = {2021},
}

@inproceedings{SinghNSJCBPR19,
  author       = {Amanpreet Singh and
                  Vivek Natarajan and
                  Meet Shah and
                  Yu Jiang and
                  Xinlei Chen and
                  Dhruv Batra and
                  Devi Parikh and
                  Marcus Rohrbach},
  title        = {Towards {VQA} Models That Can Read},
  booktitle    = {{IEEE} Conference on Computer Vision and Pattern Recognition, {CVPR}
                  2019, Long Beach, CA, USA, June 16-20, 2019},
  year         = {2019},
}

@inproceedings{SongWCWGW25,
  author       = {Dingjie Song and
                  Wenjun Wang and
                  Shunian Chen and
                  Xidong Wang and
                  Michael X. Guan and
                  Benyou Wang},
  title        = {{Less is More: {A} Simple yet Effective Token Reduction Method for Efficient Multi-modal LLMs}},
  booktitle    = {Proceedings of the 31st International Conference on Computational Linguistics, {COLING}},
  year         = {2025},
}

@misc{vicuna2023,
    title = {Vicuna: An Open-Source Chatbot Impressing GPT-4 with 90\%* ChatGPT Quality},
    url = {https://lmsys.org/blog/2023-03-30-vicuna/},
    author = {Chiang, Wei-Lin and Li, Zhuohan and Lin, Zi and Sheng, Ying and Wu, Zhanghao and Zhang, Hao and Zheng, Lianmin and Zhuang, Siyuan and Zhuang, Yonghao and Gonzalez, Joseph E. and Stoica, Ion and Xing, Eric P.},
    year = {2023}
}

@inproceedings{WenGLH025,
  author       = {Zichen Wen and
                  Yifeng Gao and
                  Weijia Li and
                  Conghui He and
                  Linfeng Zhang},
  title        = {{Token Pruning in Multimodal Large Language Models: Are We Solving the Right Problem?}},
  booktitle    = {Findings of the Association for Computational Linguistics, {ACL}},
  year         = {2025},
}

@inproceedings{WenGWZZLHZ25,
  author       = {Zichen Wen and
                  Yifeng Gao and
                  Shaobo Wang and
                  Junyuan Zhang and
                  Qintong Zhang and
                  Weijia Li and
                  Conghui He and
                  Linfeng Zhang},
  title        = {{Stop Looking for "Important Tokens" in Multimodal Language
                  Models: Duplication Matters More}},
  booktitle    = {Proceedings of the 2025 Conference on Empirical Methods in Natural Language Processing, {EMNLP}},
  year         = {2025},
}

@inproceedings{YuYLWL0WW24,
  author       = {Weihao Yu and
                  Zhengyuan Yang and
                  Linjie Li and
                  Jianfeng Wang and
                  Kevin Lin and
                  Zicheng Liu and
                  Xinchao Wang and
                  Lijuan Wang},
  title        = {{MM-Vet: Evaluating Large Multimodal Models for Integrated Capabilities}},
  booktitle    = {Forty-first International Conference on Machine Learning, {ICML}},
  year         = {2024},
}

@article{abs-2302-13971,
  author       = {Hugo Touvron and
                  Thibaut Lavril and
                  Gautier Izacard and
                  Xavier Martinet and
                  Marie{-}Anne Lachaux and
                  Timoth{\'{e}}e Lacroix and
                  Baptiste Rozi{\`{e}}re and
                  Naman Goyal and
                  Eric Hambro and
                  Faisal Azhar and
                  Aur{\'{e}}lien Rodriguez and
                  Armand Joulin and
                  Edouard Grave and
                  Guillaume Lample},
  title        = {LLaMA: Open and Efficient Foundation Language Models},
  journal      = {CoRR},
  year         = {2023},
  url          = {https://doi.org/10.48550/arXiv.2302.13971},
}

@article{abs-2310-06825,
  author       = {Albert Q. Jiang and
                  Alexandre Sablayrolles and
                  Arthur Mensch and
                  Chris Bamford and
                  Devendra Singh Chaplot and
                  Diego de Las Casas and
                  Florian Bressand and
                  Gianna Lengyel and
                  Guillaume Lample and
                  Lucile Saulnier and
                  L{\'{e}}lio Renard Lavaud and
                  Marie{-}Anne Lachaux and
                  Pierre Stock and
                  Teven Le Scao and
                  Thibaut Lavril and
                  Thomas Wang and
                  Timoth{\'{e}}e Lacroix and
                  William El Sayed},
  title        = {Mistral 7B},
  journal      = {CoRR},
  year         = {2023},
}

@article{abs-2410-17247,
  author       = {Long Xing and
                  Qidong Huang and
                  Xiaoyi Dong and
                  Jiajie Lu and
                  Pan Zhang and
                  Yuhang Zang and
                  Yuhang Cao and
                  Conghui He and
                  Jiaqi Wang and
                  Feng Wu and
                  Dahua Lin},
  title        = {{PyramidDrop: Accelerating Your Large Vision-Language Models via Pyramid Visual Redundancy Reduction}},
  journal      = {CoRR},
  year         = {2024},
}

@article{abs-2502-13923,
  author       = {Shuai Bai and
                  Keqin Chen and
                  Xuejing Liu and
                  Jialin Wang and
                  Wenbin Ge and
                  Sibo Song and
                  Kai Dang and
                  Peng Wang and
                  Shijie Wang and
                  Jun Tang and
                  Humen Zhong and
                  Yuanzhi Zhu and
                  Ming{-}Hsuan Yang and
                  Zhaohai Li and
                  Jianqiang Wan and
                  Pengfei Wang and
                  Wei Ding and
                  Zheren Fu and
                  Yiheng Xu and
                  Jiabo Ye and
                  Xi Zhang and
                  Tianbao Xie and
                  Zesen Cheng and
                  Hang Zhang and
                  Zhibo Yang and
                  Haiyang Xu and
                  Junyang Lin},
  title        = {{Qwen2.5-VL Technical Report}},
  journal      = {CoRR},
  year         = {2025},
}

@article{abs-2504-10479,
  author       = {Jinguo Zhu and
                  Weiyun Wang and
                  Zhe Chen and
                  Zhaoyang Liu and
                  Shenglong Ye and
                  Lixin Gu and
                  Hao Tian and
                  Yuchen Duan and
                  Weijie Su and
                  Jie Shao and
                  Zhangwei Gao and
                  Erfei Cui and
                  Xuehui Wang and
                  Yue Cao and
                  Yangzhou Liu and
                  Xingguang Wei and
                  Hongjie Zhang and
                  Haomin Wang and
                  Weiye Xu and
                  Hao Li and
                  Jiahao Wang and
                  Nianchen Deng and
                  Songze Li and
                  Yinan He and
                  Tan Jiang and
                  Jiapeng Luo and
                  Yi Wang and
                  Conghui He and
                  Botian Shi and
                  Xingcheng Zhang and
                  Wenqi Shao and
                  Junjun He and
                  Yingtong Xiong and
                  Wenwen Qu and
                  Peng Sun and
                  Penglong Jiao and
                  Han Lv and
                  Lijun Wu and
                  Kaipeng Zhang and
                  Huipeng Deng and
                  Jiaye Ge and
                  Kai Chen and
                  Limin Wang and
                  Min Dou and
                  Lewei Lu and
                  Xizhou Zhu and
                  Tong Lu and
                  Dahua Lin and
                  Yu Qiao and
                  Jifeng Dai and
                  Wenhai Wang},
  title        = {{InternVL3: Exploring Advanced Training and Test-Time Recipes for Open-Source Multimodal Models}},
  journal      = {CoRR},
  year         = {2025},
}

@inproceedings{zhang2025beyond,
    title={{Beyond Attention or Similarity: Maximizing Conditional Diversity for Token Pruning in {MLLM}s}},
    author={Qizhe Zhang and Mengzhen Liu and Lichen Li and Ming Lu and Yuan Zhang and Junwen Pan and Qi She and Shanghang Zhang},
    booktitle={The Thirty-ninth Annual Conference on Neural Information Processing Systems, {NeurIPS}},
    year={2025},
}

@article{abs-2407-07895,
  author       = {Feng Li and
                  Renrui Zhang and
                  Hao Zhang and
                  Yuanhan Zhang and
                  Bo Li and
                  Wei Li and
                  Zejun Ma and
                  Chunyuan Li},
  title        = {LLaVA-NeXT-Interleave: Tackling Multi-image, Video, and 3D in Large
                  Multimodal Models},
  journal      = {CoRR},
  year         = {2024},
}

@article{abs-2509-24837,
  author       = {Youngeun Kim and
                  Youjia Zhang and
                  Huiling Liu and
                  Aecheon Jung and
                  Sunwoo Lee and
                  Sungeun Hong},
  title        = {{Training-Free Token Pruning via Zeroth-Order Gradient Estimation in Vision-Language Models}},
  journal      = {CoRR},
  year         = {2025},
}

@inproceedings{0020FMZ0CGONKZ25,
  author       = {Yuan Zhang and
                  Chun{-}Kai Fan and
                  Junpeng Ma and
                  Wenzhao Zheng and
                  Tao Huang and
                  Kuan Cheng and
                  Denis A. Gudovskiy and
                  Tomoyuki Okuno and
                  Yohei Nakata and
                  Kurt Keutzer and
                  Shanghang Zhang},
  title        = {{SparseVLM: Visual Token Sparsification for Efficient Vision-Language Model Inference}},
  booktitle    = {Forty-second International Conference on Machine Learning, {ICML}},
  year         = {2025},
}

@article{0080ZGZ00ZZL0L25,
  author       = {Bo Li and
                  Yuanhan Zhang and
                  Dong Guo and
                  Renrui Zhang and
                  Feng Li and
                  Hao Zhang and
                  Kaichen Zhang and
                  Peiyuan Zhang and
                  Yanwei Li and
                  Ziwei Liu and
                  Chunyuan Li},
  title        = {{LLaVA-OneVision: Easy Visual Task Transfer}},
  journal      = {Trans. Mach. Learn. Res.},
  year         = {2025},
}

@article{SHITOV202152,
title = {{Column subset selection is NP-complete}},
journal = {Linear Algebra and its Applications},
volume = {610},
pages = {52-58},
year = {2021},
author = {Yaroslav Shitov},
}

@inproceedings{FazlyabRHMP19,
  author       = {Mahyar Fazlyab and
                  Alexander Robey and
                  Hamed Hassani and
                  Manfred Morari and
                  George J. Pappas},
  title        = {{Efficient and Accurate Estimation of Lipschitz Constants for Deep
                  Neural Networks}},
  booktitle    = {Annual Conference on Neural Information Processing Systems, {NeurIPS}},
  pages        = {11423--11434},
  year         = {2019},
}

\clearpage
\nolinenumbers

\setcounter{table}{0}
\setcounter{figure}{0}

\renewcommand{\thetable}{A\arabic{table}}
\renewcommand{\thefigure}{A\arabic{figure}}

\appendix

\section*{\centering Supplementary Material}
\section*{\centering Moving Beyond Diversity: Visual Token Pruning as Subspace Reconstruction for Efficient VLMs}

\section{Appendix} 
In this appendix, we first present detailed descriptions of the experimental setup, including the models, benchmarks, and implementation details.
We then provide additional experimental results, sensitivity analysis, and visualizations.

\subsection{Details of Experimental Setup}

\smalltitle{Model Details.}
\begin{itemize}
\renewcommand\labelitemi{$\bullet$}
    \item \textbf{LLaVA-1.5 \cite{LiuLLL24}.} A widely used open-source vision-language model that processes $336 \times 336$ images and produces 576 visual tokens per image. The architecture consists of a CLIP ViT-L/14 vision encoder \cite{RadfordKHRGASAM21}, a two-layer MLP multimodal projector with GELU activation, and a large language model based on Vicuna \cite{vicuna2023}.
    \item \textbf{LLaVA-NeXT \cite{liu2024llavanext}.} An improved version of LLaVA designed for high-resolution visual understanding. In our experiments, we fix the resolution of input images to $672 \times 672$ resolution, resulting in 2880 visual tokens per image. The architecture largely follows that of LLaVA-1.5 \cite{LiuLLL24}.
    \item \textbf{Qwen2.5-VL \cite{abs-2502-13923}.}
    A recent model based on the Qwen-VL architecture, designed to improve multimodal perception and reasoning.  It employs a Naive Dynamic Resolution (NDR) strategy to process images of arbitrary aspect ratios, dynamically generating a variable number of visual tokens. The architecture consists of a ViT-based vision encoder with Multimodal Rotary Positional Embedding (MRoPE), a vision-language merger, and the Qwen2.5 large language model. This design enables the model to capture spatial and temporal relationships while supporting long-context sequences of up to 128K tokens.
\end{itemize}

\smalltitle{Benchmark Details.}
\begin{itemize}
\renewcommand\labelitemi{$\bullet$}
    \item \textbf{VQAv2 \cite{GoyalKSBP17}.} A balanced visual question answering benchmark designed to reduce language priors by pairing visually similar images that correspond to different answers for the same question. The dataset contains about 1.1M image–question pairs from approximately 200K MS COCO images, each annotated with ten human-provided answers.
    \item \textbf{GQA \cite{HudsonM19}.} A large-scale benchmark for compositional visual reasoning designed to mitigate strong statistical biases in earlier VQA datasets. It contains over 22M questions over 113K images from Visual Genome scene graphs, where each question is associated with a structured semantic representation.
    \item \textbf{SQA \cite{LuMX0CZTCK22}.} A large-scale multimodal benchmark designed to evaluate multi-hop reasoning across diverse scientific domains. It contains 21K questions spanning natural science, social science, and language science, many of which are annotated with grounded lectures and explanations. In our experiments, we use the SQA-IMG subset, which contains questions paired with images for multimodal reasoning evaluation.
    \item \textbf{TextVQA \cite{SinghNSJCBPR19}.} A visual question answering benchmark designed to evaluate a model’s ability to read and reason about text embedded in real-world images. It contains 45K human-generated questions over 28K images from Open Images v3, requiring both visual understanding and optical character recognition.
    \item \textbf{POPE \cite{LiDZWZW23}.} A benchmark designed to evaluate object hallucination in vision-language models. It formulates hallucination detection as a binary question answering task about object presence under three evaluation settings: random, popular, and adversarial.
    \item \textbf{MME \cite{fu2025mme}.} A comprehensive benchmark designed to evaluate the perception and cognition abilities of multimodal large language models. It uses concise yes/no questions and includes paired questions with opposite answers across 14 subtasks covering visual perception and cognitive reasoning.
    \item \textbf{MMBench \cite{LiuDZLZZYWHLCL24}.} A comprehensive bilingual benchmark for evaluating fine-grained multimodal perception and reasoning abilities of vision-language models. It contains 3,217 carefully curated multiple-choice questions in English and Chinese across 20 ability dimensions and adopts the CircularEval strategy to reduce position bias.
    \item \textbf{MM-Vet \cite{YuYLWL0WW24}.} A benchmark designed to evaluate large multimodal models on complex tasks that require integrating multiple vision-language capabilities. It contains 218 open-ended questions over 200 curated images and assesses combinations of six core capabilities, including recognition, OCR, knowledge, language generation, spatial reasoning, and math. For evaluation, we use GPT-4.1 as the judge to score model responses.
\end{itemize}

\smalltitle{Implementation Details.}
All experiments are conducted on 4$\times$NVIDIA A6000 GPUs with a batch size of 1. 
SPARE is fully training-free and applied at inference time. 
Specifically, visual token pruning is performed after the multimodal projector and before the visual tokens are passed to the LLM.

For LLaVA, we use the official implementations\footnote{\url{https://github.com/haotian-liu/LLaVA}}, while experiments on Qwen2.5-VL are conducted using the \texttt{lmms-eval}\footnote{\url{https://github.com/EvolvingLMMs-Lab/lmms-eval}} evaluation framework. 
We evaluate three pruning ratios corresponding to 77.8\%, 88.9\%, and 94.4\% token reduction for the LLaVA models, and report the overall performance using the average relative accuracy across benchmarks.
For Qwen2.5-VL, we evaluate settings that retain 20\% and 10\% of the visual tokens for each image, reflecting its dynamic tokenization rather than using a fixed token budget.

\subsection{Additional Experiments and Analyses}

\smalltitle{Performance Comparison on LLaVA-NeXT-13B.} 
We further evaluate SPARE on a larger model, LLaVA-NeXT-13B \cite{liu2024llavanext}, to examine the generalization of our method. As shown in Table~\ref{tab1_llava_next_appendix}, SPARE achieves strong overall performance across pruning ratios. Similar to the observations on LLaVA-NeXT-7B \cite{liu2024llavanext}, the improvement becomes less pronounced when retaining 22.2\% of tokens (Retain 640 tokens), likely because the large number of preserved tokens reduces the impact of token selection in this high-resolution setting. Nevertheless, under extreme pruning (\ie, 94.4\% token reduction), SPARE again demonstrates strong robustness, preserving about 96\% of the original performance.

\begin{table}[t]
\caption{\textbf{Performance Comparison on LLaVA-NeXT-13B.}}
\centering
\small
\setlength{\tabcolsep}{3.8pt}
\renewcommand{\arraystretch}{1.05}

\resizebox{\linewidth}{!}{%
\begin{tabular}{l|ccccccccc|c}
\toprule
\textbf{Method} &
\textbf{VQA$^{\mathrm{V2}}$} & \textbf{GQA} & \textbf{SQA$^{\mathrm{Img}}$} & \textbf{VQA$^{\mathrm{Text}}$} &
\textbf{POPE} & \textbf{MME} & \textbf{MMB$^{\mathrm{EN}}$} & \textbf{MMB$^{\mathrm{CN}}$} & \textbf{MMVet} & \textbf{Rel.\ Acc} \\

\rowcolor{gray!15}
\multicolumn{11}{c}{\textit{Retain 2880 tokens (100\%)}}\\
LLaVA-NeXT-13B
& 82.3 & 64.4 & 73.1 & 63.2 & 85.3 & 1539.5 & 68.5 & 61.2 & 45.0 & 100.0\% \\

\rowcolor{gray!15}
\multicolumn{11}{c}{\textit{Retain 640 tokens \textcolor{green!70!black}{$(\downarrow 77.8\%)$}}}\\
DART ({\color{black!60}EMNLP 25})
& \textbf{81.8} & \textbf{64.4} & \textbf{73.7} & \textbf{64.7} & 84.7 & 1558.1 & 68.6 & \textbf{64.0} & \textbf{44.3} & \textbf{\textcolor{red}{100.7}} \\
CDPruner ({\color{black!60}NeurIPS 25})
& 81.0 & 64.0 & 71.9 & 61.0 & \textbf{87.6} & 1539.2 & \textbf{69.1} & 62.5 & 39.0 & 98.3 \\
\textbf{SPARE (Ours)}
& 80.7 & 64.0 & 71.4 & 61.1 & 87.3 & \textbf{1579.6} & 68.6 & 63.0 & 41.3 & 99.1 \\

\rowcolor{gray!15}
\multicolumn{11}{c}{\textit{Retain 320 tokens \textcolor{green!70!black}{$(\downarrow 88.9\%)$}}}\\
DART ({\color{black!60}EMNLP 25})
& \textbf{79.7} & 62.2 & \textbf{72.1} & 58.8 & 81.7 & \textbf{1528.6} & 66.6 & 61.9 & 39.9 & 96.3 \\
CDPruner ({\color{black!60}NeurIPS 25})
& 79.5 & 63.1 & 71.1 & 58.8 & \textbf{87.5} & 1496.7 & 66.5 & 62.0 & \textbf{40.8} & 97.1 \\
\textbf{SPARE (Ours)}
& 79.0 & \textbf{63.4} & 70.7 & \textbf{59.3} & 87.2 & \textbf{1528.6} & \textbf{67.0} & \textbf{62.5} & 39.5 & \textbf{\textcolor{red}{97.2}} \\

\rowcolor{gray!15}
\multicolumn{11}{c}{\textit{Retain 160 tokens \textcolor{green!70!black}{$(\downarrow 94.4\%)$}}}\\
DART ({\color{black!60}EMNLP 25})
& 75.1 & 58.1 & \textbf{71.3} & 49.6 & 75.7 & 1448.4 & 63.6 & 57.7 & 34.8 & 89.4 \\
CDPruner ({\color{black!60}NeurIPS 25})
& \textbf{77.8} & \textbf{62.1} & 70.4 & 56.7 & \textbf{88.5} & 1484.7 & 66.2 & 60.4 & \textbf{38.0} & 95.2 \\
\textbf{SPARE (Ours)}
& 77.3 & 62.0 & 70.8 & \textbf{57.9} & 87.6 & \textbf{1532.3} & \textbf{67.3} & \textbf{61.8} & 36.2 & \textbf{\textcolor{red}{95.6}} \\

\bottomrule
\end{tabular}
}
\label{tab1_llava_next_appendix}
\end{table}

\begin{table}[t]
\caption{\textbf{Detailed Results on the Multi-Skill Benchmark.}}
\centering
\small
\setlength{\tabcolsep}{3pt}
\renewcommand{\arraystretch}{1.05}

\resizebox{\linewidth}{!}{%
\begin{tabular}{l|cc|ccc|ccccc|cc|c}
\toprule
\textbf{Method} &
\multicolumn{2}{c|}{\textbf{1-Skill}} &
\multicolumn{3}{c|}{\textbf{2-Skill}} &
\multicolumn{5}{c|}{\textbf{3-Skill}} &
\multicolumn{2}{c|}{\textbf{4-Skill}} &
\textbf{Total} \\

& \textit{r} & \textit{o}
& \textit{rk} & \textit{rs} & \textit{ro}
& \textit{rkg} & \textit{os} & \textit{osm} & \textit{ros} & \textit{ogs}
& \textit{rogs} & \textit{rokg}
& \\

\midrule

\rowcolor{gray!15}
\multicolumn{14}{c}{\textit{Retain 32 tokens \textcolor{green!70!black}{$(\downarrow 94.4\%)$}}} \\

CDPruner ({\color{black!60}NeurIPS 25})
& \textbf{60.8} & 40.0
& \textbf{16.7} & \textbf{56.7} & 50.0
& 19.8 & 15.4 & 19.3 & 0.0 & 0.0
& 37.5 & 16.2
& 27.9 \\

\textbf{SPARE (Ours)}
& 48.6 & \textbf{40.8}
& \textbf{16.7} & 50.0 & \textbf{70.0}
& \textbf{26.8} & \textbf{24.6} & \textbf{27.1} & \textbf{24.3} & \textbf{15.0}
& \textbf{50.0} & \textbf{18.8}
& \textbf{31.4} \\

\bottomrule
\end{tabular}%
}

\vspace{2pt}

\label{tab2_mmvet_appendix}
\end{table}

\smalltitle{Performance Comparison on Multi-Skill Benchmark.}
We provide a detailed comparison of compositional benchmark across internal tasks, grouped by the number of required reasoning skills.
Specifically, we report the results of LLaVA-1.5-7B \cite{LiuLLL24} on the MM-Vet benchmark \cite{YuYLWL0WW24} with 32 retained visual tokens.
In table \ref{tab2_mmvet_appendix}, the symbols \textit{r}, \textit{o}, \textit{k}, \textit{g}, \textit{s}, and \textit{m} denote recognition, OCR, knowledge, generation, spatial reasoning and math, respectively. For brevity, we do not report categories where both SPARE and CDPruner \cite{zhang2025beyond} achieve zero performance.
As shown in Table \ref{tab2_mmvet_appendix}, while SPARE does not show a clear advantage on single-skill tasks, it consistently outperforms CDPruner \cite{zhang2025beyond} on tasks requiring multiple reasoning skills, suggesting that SPARE is particularly effective for compositional reasoning.

\smalltitle{Sensitivity Analysis of Anti-Relevance Measurement.}
To analyze the impact of different formulations for deriving anti-relevance, we evaluate several design choices. Specifically, we first compute the cosine similarity between image and text embeddings to obtain the relevance scores. We then consider two aggregation strategies over text tokens, taking either the mean or the maximum value. Afterward, we apply different normalization schemes, including the commonly used min–max normalization \cite{zhang2025beyond, abs-2509-24837} and our formulation based on z-score normalization followed by a softplus function.
We conduct the analysis on LLaVA-NeXT-7B \cite{liu2024llavanext} with 160 retained visual tokens (\ie, 94.4\% token reduction). 
As shown in Table~\ref{tab3_antirelevancemetric_appendix}, although the performance differences are modest, mean aggregation with z-score normalization followed by a softplus function yields more stable results across benchmarks.
Based on these observations, we adopt this metric for computing anti-relevance in SPARE.

\begin{figure}[t]
    \centering
    \includegraphics[width=1.0\linewidth]{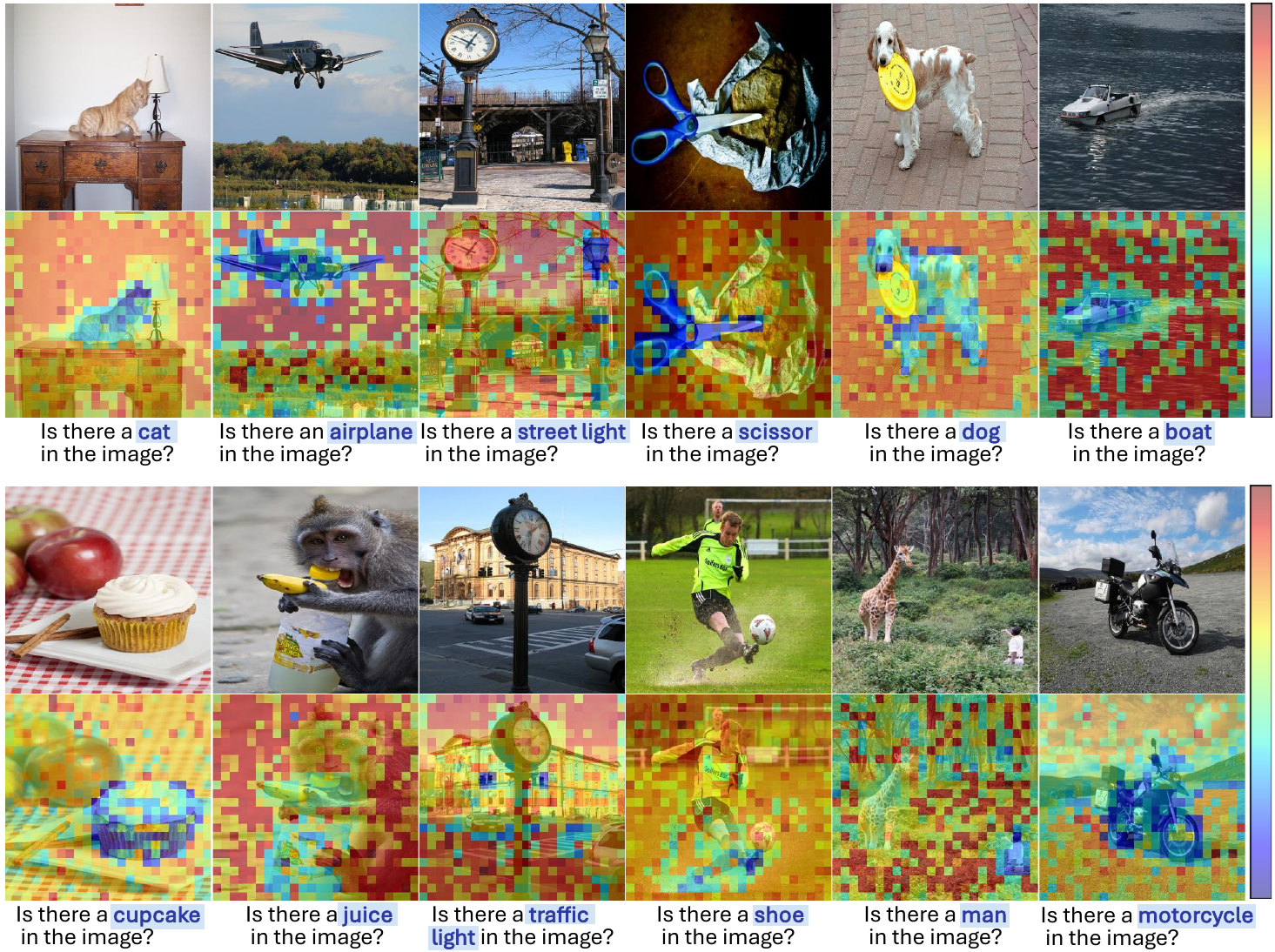}
        \vspace{-1mm}
    \caption{\textbf{Visualizations of Image-Text Relevance on the POPE Benchmark \cite{LiDZWZW23}.} Red denotes higher relevance, while blue denotes lower relevance.}
    \label{fig1_vis_appendix}
    \vspace{-2mm}
\end{figure}

\smalltitle{Visualizations.}
To further investigate the behavior of SPARE, we provide additional visualizations. 
In Fig. \ref{fig1_vis_appendix}, we visualize the image-text relevance scores to examine their distribution across visual tokens.
As shown, objects that are semantically related to the textual query often exhibit relatively low relevance scores, while some background regions or visually dominant areas receive higher relevance values. 
This observation motivates the use of our anti-relevance guidance, which enables token selection to incorporate contextual information beyond image-only selection that ignores the textual query.

\begin{figure}[t]
    \centering
    \includegraphics[width=0.85\linewidth]{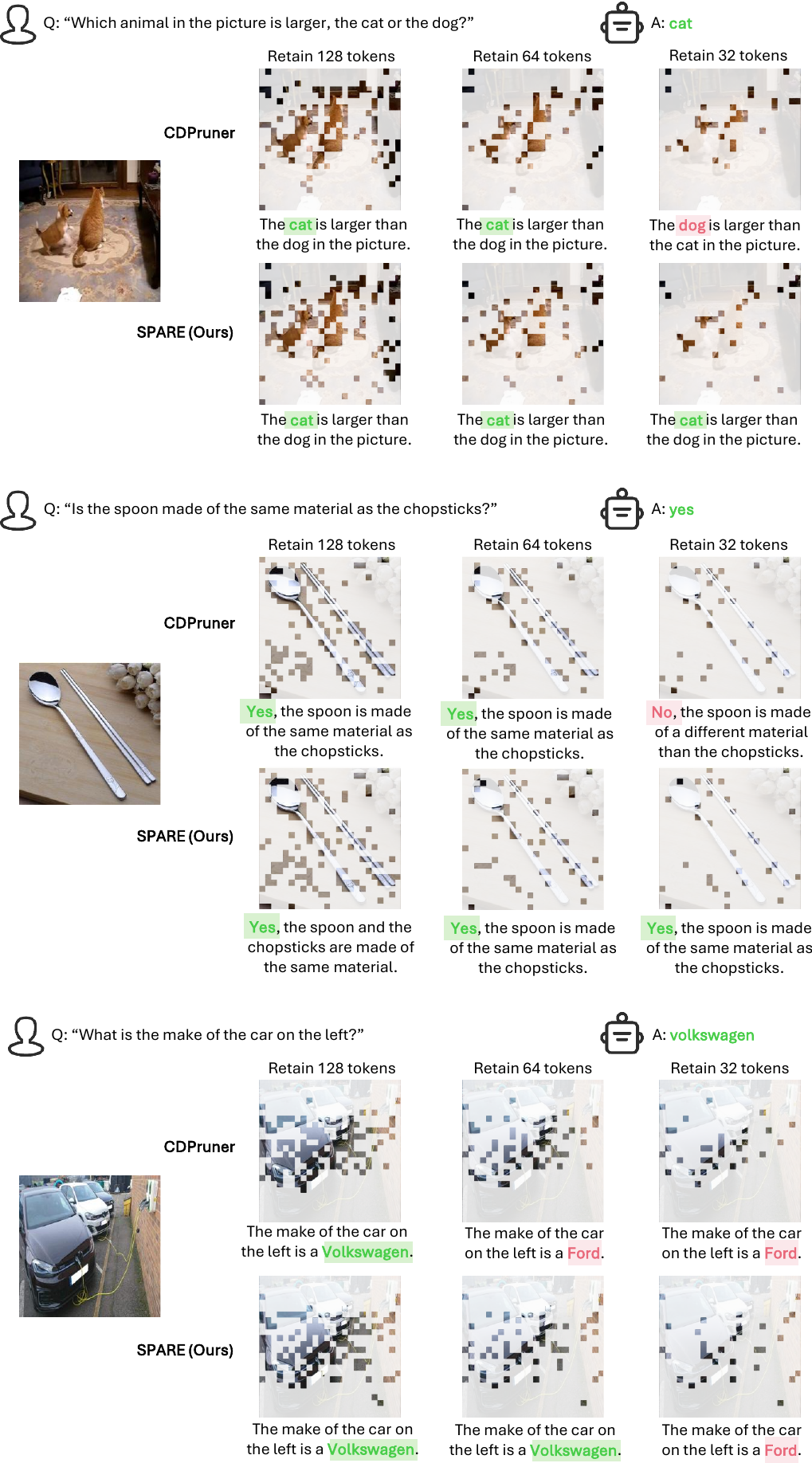}
        \vspace{-1mm}
    \caption{\textbf{Qualitative Comparison.} Visualizations of retained tokens under different token retention budgets, comparing SPARE with CDPruner \cite{zhang2025beyond}.}
    \label{fig2_vis_appendix}
    \vspace{-2mm}
\end{figure}

In Fig. \ref{fig2_vis_appendix}, we further present visualization results under different token retention budgets and compare our method with CDPruner \cite{zhang2025beyond}. 
As the number of retained tokens decreases, a difference between the two approaches becomes evident, with our method maintaining more informative visual evidence under extremely low token budgets, highlighting the importance of reconstruction-based token selection.

\begin{table}[t]
\caption{\textbf{Analysis of Anti-Relevance Measurement Strategies.}}
\centering
\small
\setlength{\tabcolsep}{4pt}
\renewcommand{\arraystretch}{1.05}

\resizebox{0.7\linewidth}{!}{%
\begin{tabular}{lc|cccc}
\toprule
\textbf{Metric} &  & \textbf{GQA} & \textbf{SQA} & \textbf{POPE} & \textbf{MME} \\
\midrule

\rowcolor{gray!15}
\multicolumn{6}{c}{\textit{Retain 2880 tokens}}\\
LLaVA-NeXT-7B &  & 62.5 & 67.5 & 86.8 & 1511.8 \\

\rowcolor{gray!15}
\multicolumn{6}{c}{\textit{Retain 160 tokens \textcolor{green!70!black}{$(\downarrow 94.4\%)$}}}\\

\multirow{2}{*}{min-max normalization}
& mean & 60.8 & 67.1 & 86.0 & 1400.8 \\
& max & 60.8 & \textbf{67.6} & 86.0 & 1400.8 \\

\addlinespace[3pt]

\multirow{2}{*}{z-score + softplus}
& mean & \textbf{61.0} & 67.5 & \textbf{86.4} & \textbf{1409.9} \\
& max & 60.7 & 67.5 & \textbf{86.4} & \textbf{1409.9} \\

\bottomrule
\end{tabular}
}

\label{tab3_antirelevancemetric_appendix}
\end{table}

\end{document}